\documentclass[preprint,11pt,authoryear]{elsarticle}
\usepackage[utf8]{inputenc}
\usepackage[T1]{fontenc}   

\usepackage{geometry}
\usepackage{setspace}

\usepackage{amsmath,amsfonts,amssymb,amsthm}
\usepackage{bm,physics,mathtools}

\usepackage{graphicx}            
\usepackage{booktabs}            
\usepackage{multirow}            
\usepackage{float}               
\usepackage{tikz}               

\usepackage{natbib}

\usepackage{enumitem}            
\usepackage{comment}            
\usepackage[colorlinks=true, linkcolor=blue, urlcolor=blue, citecolor=blue]{hyperref}

\journal{Nuclear Physics B}          %

\begin{document}

\begin{frontmatter}



\title{Physics-Embedded Gaussian Process for Traffic State Estimation} 

\author{Yanlin Chen} 
\ead{yanlinc@uw.edu}

\author{Kehua Chen\corref{cor1}} 
\ead{zeonchen@uw.edu}

\author{Yinhai Wang\corref{cor2}}
\ead{yinhai@uw.edu}

\cortext[cor1]{Corresponding author}
\cortext[cor2]{Corresponding author}

\affiliation{organization={Department of Civil and Environmental Engineering, University of Washington},
            addressline={}, 
            city={Seattle},
            postcode={98195-2700}, 
            state={WA},
            country={USA}}

\begin{abstract}
Traffic state estimation (TSE) becomes challenging when probe‐vehicle penetration is low and observations are spatially sparse. Pure data-driven methods lack physical explanations and have poor generalization when observed data is sparse. In contrast, physical models have difficulty integrating uncertainties and capturing the real complexity of traffic. To bridge this gap, recent studies have explored combining them by embedding physical structure into 
Gaussian process. These approaches typically introduce the governing equations as soft constraints through pseudo-observations, enabling the integration of model structure within a variational framework. However, these methods rely heavily on penalty tuning and lack principled uncertainty calibration, which makes them sensitive to model mis-specification. In this work, we address these limitations by presenting a novel Physics-Embedded Gaussian Process (PEGP), designed to integrate domain knowledge with data-driven methods in traffic state estimation. Specifically, we design two multi-output kernels informed by classic traffic flow models (i.e. Aw–Rascle–Zhang (ARZ) model and Lighthill-Whitham-Richards (LWR) Model), constructed via the explicit application of the linearized differential operator. Experiments on HighD, NGSIM show consistent improvements over non-physics baselines. PEGP–ARZ proves more reliable under sparse observation, while PEGP–LWR achieves lower errors with denser observation. Ablation study further reveals that PEGP--ARZ residuals align closely with physics and yield calibrated, interpretable uncertainty, whereas PEGP--LWR residuals are more orthogonal and produce nearly constant variance fields. This PEGP framework combines physical priors, uncertainty quantification, which can provide reliable support for TSE. The code for our implementation of the PEGP framework is available at our public GitHub repository: \url{https://github.com/Eddiech-coder/PEGP}.
\end{abstract}

\end{frontmatter}
\section{Introduction}
Amid urbanization and rising traffic demand, traffic state estimation (TSE) faces higher performance requirements, posing new challenges for traffic signal control, trajectory planning, and congestion management in modern intelligent transportation systems \citep{agarwal2022sensing}.
The observational data used in traffic state estimation can be divided into two major categories \citep{bekiaris2016highway}. One is the speed, flow or density information provided by fixed detection equipment (such as such as loop and radar detector). The other category involves floating perception data derived from probe vehicles or Connected Vehicles (CVs), including GPS trajectories and travel times. 

Given these observational data, the TSE method is mainly divided into two types: the first-principle method and the data-driven method \citep{antoniou2013dynamic}. The first-principle method, constructed based on fundamental physical laws such as mass conservation and momentum conservation of traffic flow, has a clear model structure and explicit parameter definitions. Therefore, it has good interpretability and robust extrapolation capabilities \citep{seo2017traffic}. In contrast, the data-driven approach can flexibly capture spatio-temporal patterns on road where large-scale historical and real-time data are available and traffic patterns are complex. This thus demonstrates higher prediction accuracy and adaptability. The existing data-driven methods mainly include statistical learning methods (such as Kalman filter and particle filter) \citep{wang2005real}, time series analysis (such as ARIMA and state space model) \citep{stathopoulos2003multivariate}, deep and machine learning methods. Among those machine learning methods, Gaussian Process (GP), as a nonparametric Bayesian regression method, is widely used in the field of traffic state estimation \citep{wu2024traffic, liu2023gaussian, wang2023low} because it can naturally describe the spatial-temporal correlation and provide uncertainty quantification. At the same time, it can handle multi-scale data fusion, anomaly detection and missing data interpolation \citep{chen2015gaussian}. This is of great significance for traffic management decision-making and dealing with common sparse observations in traffic systems.

However, GPs fail to incorporate the aforementioned fundamental physical laws within their model structure, leading to a lack of physical consistency in model predictions, and non-physical phenomena such as negative density or unreasonable velocity may occur. This defect is particularly prominent when observations are sparse or extrapolation is performed. Additionally, such a weakness compromises the model's generalization ability, rendering it more vulnerable to severe physical errors.

Building upon the preceding analysis, we propose a novel TSE method called Physics-Embedded Gaussian Process (PEGP) that seeks to harmoniously integrate the prior of physical models with the adaptability of machine learning approaches. We start by deriving linear operators from the Partial Differential Equation (PDE) structures of the LWR and ARZ traffic flow models
For PEGP--LWR, choose a typical traffic state, write the states as a baseline plus small perturbations, and keep first–order terms. The nonlinear conservation law simplifies to a linear advection equation. The propagation speed is determined by the local wave speed, which is given by the derivative of the flow–density relation at the chosen state. This linear form describes how disturbances travel along characteristic directions and can also accommodate simple source or coupling effects if needed. For PEGP--ARZ, select a chosen steady state, introduce small perturbations in speed and density or in the corresponding Riemann invariants, rewrite the model in Riemann coordinates, and linearize to obtain two approximately decoupled first–order dynamics, one per characteristic branch, with relaxation appearing as a constant relaxation term. The coefficients are set by the traffic conditions at the chosen state, including the local speed, density, and how pressure varies with density. In both cases, the linear operators are derived by separating the temporal and spatial derivatives in the linearized equations, which define the characteristic directions of propagation together with any constant relaxation terms.
Subsequently, these operators are applied to the squared exponential kernel to develop a physics-embedded kernel that explicitly incorporates physical knowledge. Thus, the proposed multi-output GP model seamlessly embeds the physical prior into the kernel structure, unifying the benefits of data-driven and physical methods. We further evaluate the framework on HighD and NGSIM, where physics-embedded GPs consistently outperform non-physics baselines. PEGP–ARZ offers greater robustness under sparse observation and interpretable uncertainty, while PEGP–LWR attains lower errors with denser observation. Taken together, the results demonstrate that embedding physics priors into Gaussian processes enhances both the accuracy and interpretability of traffic state estimation.
To sum up, the main contributions of this work are as follows:
\begin{itemize}
    \item We derive physics-embedded GP kernels from linearized ARZ and LWR operators, yielding closed-form cross-covariances for $(\rho,u)$ and a positive definite, physically dimensioned construction that captures asymmetric wave propagation.
    \item Across NGSIM and HighD datasets, both PEGP models consistently outperform non-PEGP baselines. PEGP–LWR achieves higher reconstruction accuracy with denser observation, while PEGP–ARZ provides calibrated, interpretable uncertainty and greater diagnostic value under sparse observation.
    \item We introduce a physics–residual decomposition with aligned and energy shares, joint ratios, the Centered Kernel Alignment (CKA), and principal angles, which provides a systematic account of why PEGP–ARZ is more reliable under sparse observation, whereas PEGP–LWR attains lower errors under higher observation in the HighD dateset, particularly in congested regimes.
\end{itemize}

The remaining paper is organized as follows: Section~2 situates recent work within physical, purely data-driven, and emerging physics-informed approaches, highlighting their connections and contrasts; Section~3 presents the operator linearization, covariance kernel and the multi-output SVGP inference scheme; Sections~4 present the experimental evaluation. We first conduct case studies on HighD and NGSIM, covering reconstructions under varying penetration rates and a four-loop-detector experiment. We then perform ablation analyses of physics–residual interactions using aligned and energy shares, joint ratios, CKA, and principal angles, and extend the evaluation to uncertainty quantification. Finally, Section 5 concludes the paper with a summary of key findings and future research directions.

\section{Literature review}
Over the past several decades, the TSE has advanced from first-principle methods to a data-driven paradigm, ultimately evolving into a mixed modeling framework that integrates the two. This section systematically examines three core methodological branches: physical methods, pure data-driven methods, and the emerging physics-informed methods.

\subsection{Physical Models for Traffic State Estimation}
Within the field of traffic state estimation, macroscopic traffic flow models have attracted much attention for their well-defined structure and strong interpretability in representing traffic behavior. The LWR model proposed by \citet{lighthill1955kinematic} and \citet{richards1956shock} serves as the cornerstone of traffic flow theory, depicting the variation laws of traffic density in both time and space dimensions through first-order PDEs. However, its inherent assumption of traffic flow equilibrium limits the ability to accurately represent the non-equilibrium dynamics frequently encountered in real-world traffic. To address its shortcomings, Payne-Whitham (PW) \citep{payne1971model, whitham2011linear} first proposed a second-order model framework, followed by \citet{aw2000resurrection} and \citet{ zhang2002non}, who developed an improved second-order traffic flow model (i.e. ARZ model). These approaches establish a more comprehensive model by incorporating the driver's anticipated velocity as an additional state variable. With the development of sensor networks, combining physical models with data assimilation technology has become a key path to improve the accuracy of traffic state estimation. \citet{wang2005real} proposed a highway traffic state estimation method based on extended Kalman filtering, achieving synchronous online estimation of density, velocity and model parameters.

\subsection{Data‐Driven Methods for Traffic State Estimation}
In contrast to physical models, data-driven approaches bypass the explicit modeling of traffic flow mechanisms and instead construct models directly from observed data. Early explorations primarily focused on statistical time-series methods. For instance, the autoregressive integrated moving average (ARIMA) model \citep{zhong2004estimation} and its seasonal variant (SARIMA) \citep{shekhar2007adaptive} have been widely applied for short-term estimation of traffic flow and speed, as they capture the inherent linear dynamics of time series through autocorrelation. Beyond statistical approaches, classical machine learning methods have also been adopted. The K-nearest neighbor (K-NN) algorithm \citep{tak2016data} captures local nonlinear fluctuations by retrieving the most similar historical observations, while random forest \citep{leshem2007traffic} constructs ensembles of decision trees with random sub-features to effectively mine higher-order nonlinear relationships and improve predictive robustness.

However, the above methods lack the ability to capture spatio-temporal couplings and complex nonlinear dependencies, and they cannot provide uncertainty quantification, so they face significant challenges in modeling highly complex spatio-temporal relationships. This has motivated the adoption of GPs, which have received extensive attention in traffic state estimation. The key advantages of GPs are their ability to capture high-dimensional nonlinear relationships beyond the reach of physical models and their probabilistic framework that enables uncertainty quantification. In addition to their general strengths, GPs have been extended in various ways to better address the requirements of traffic state estimation. 
In terms of model stability, \citet{liu2022gaussian} introduced the "hat kernel" to address the hyperparameter optimization problem of GPR in TSE and derived the lower bound to avoid the deformation and overfitting problems that may occur in traditional RBF kernels. Subsequently, \citet{lei2024unraveling} constructed a random base graph based on Sparse GPR and verified that the zero-mean SGPR can accurately fit both velocity and flow. In addition, they innovatively incorporated empirical knowledge into the learning process by employing the empirical deterministic fundamental graph as the mean function. To tackle the challenges of multi-source heterogeneity and severe data sparsity in large-scale traffic networks, \citet{wang2023low} proposed a spatio-temporal Hankel low-rank tensor completion framework, which proved effective in handling data loss. The prediction framework constructed by \citet{varga2023data} based on the Kriging interpolation has opened up a new path for traditional statistical interpolation in urban traffic prediction.
In response to the anisotropy of traffic wave propagation, the latest research \citep{wu2024traffic} designed a direction-sensitive kernel through a rotation matrix, enabling GPR to better accurately reconstruct the velocity field even at extremely low penetration rates.

\subsection{Physics-Informed Data-Driven Methods for Traffic State Estimation}
Although physical models have high interpretability, their simplified assumptions make it difficult for them to capture the real complexity of traffic. Conversely, data-driven models have strong fitting capabilities but behave as black boxes, sacrificing interpretability and sometimes yielding non‑physical insights.

Among those implementation paths, physics-informed data-driven methods have achieved integration of first principle and data-driven models through the integration of physical knowledge into the model structure. One approach is to minimize both the data fitting error and the physical law residuals simultaneously by combining the loss function.
\citet{mo2021physics} employed multiple car following models in the physics-informed deep learning car-following model (PIDL-CF) framework and introduced a residual trade-off mechanism to achieve precise modeling of micro-vehicle behavior. \citet{zhao2022integrating} and \citet{zhao2023observer} innovatively developed the Observer-Informed Deep Learning (OIDL) framework, which incorporates residuals as online regularization terms, leading to substantially improved adaptability in dynamic traffic environments. Another combination approach is to directly integrate fundamental diagram (FD) functions into the computational graph, using PDE or FD residuals as soft constraints to participate in the backpropagation process. \citet{shi2021physics} replaced the FD function with a neural surrogate and included conservation law residuals in the loss to jointly learn the traffic state and the FD parameters. Similarly, \citet{zhang2024physics} treated the free-flow speed and congestion density from the LWR model based on Greenshields FD as trainable weights, inferring key parameters from sparse flow-density observations and precisely reconstructing traffic congestion waves.

Another physics-informed approach aims to integrate physical models with GP called physics-informed Gaussian Process (PIGP). The current PIGP methods mainly rely on the Latent Force Model (LFM) \citep{alvarez2013linear, pmlr-v5-alvarez09a}, integrating physical operators into the GP in the form of implicit kernels. \citet{yuan2021macroscopic} and \citet{yuan2021traffic} constructed an augmented latent force model (ALFM) form of shadow GP for single-output macroscopic continuous PDE (LWR/PW/ARZ), and a soft constraint through pseudo-observations in the ELBO, using variational inference processing. However, the ALFM formulation of shadow Gaussian processes introduces the governing PDE only as a soft constraint through pseudo-observations in the ELBO, so model performance and uncertainty heavily depend on the tuning of penalty weights ~$\lambda$. Although multi-output extensions exist, cross-output dependencies are not operator-consistent but instead stem from shared latent forces or convolution kernels, which cannot guarantee conservation relations or characteristic flow. As a result, the method remains sensitive to model mis-specification, lacks principled diagnostics to separate physics from residual effects, and provides no inherent mechanism for calibrated uncertainty.

\section{Methodology}
To overcome the respective limitations of pure physical models and pure data-driven methods under sparse observation conditions, this section proposes a modeling framework that integrates the advantages of both. Our goal is to construct a joint prior that strictly adheres to the conservation of mass and momentum balance of traffic flow while also possessing the flexibility of Gaussian processes.
In this section, we introduce and derive PEGP-ARZ, a physical prior embedding method grounded in the ARZ model. The details of PEGP-LWR can be found in~\ref{subsec:lwr_kernel_cn}. Specifically, we (i) introduce Riemann
invariants; (ii) linearize around a user–chosen equilibrium; (iii)
freeze the characteristic speeds; and (iv) act the resulting constant-coefficient
matrix operator on a base SE kernel. The outcome is a closed-form $2\times2$ block covariance that embeds both the
mass-conservation and momentum-balance equations.  
The full pipeline transforms raw spatio-temporal measurements into joint probabilistic predictions of density and velocity with quantified uncertainty while respecting the ARZ dynamics. 

\subsection{ARZ Model and Its Linearization}

The relaxation form of the ARZ model is \citep{aw2000resurrection, zhang2002non}:

\begin{equation}\label{eq:arz}
{%
\begin{aligned}
\partial_t\rho + \partial_x(\rho\,v) &= 0, \\[4pt]
\partial_t\!\bigl(v + P(\rho)\bigr) + v\,\partial_x\!\bigl(v + P(\rho)\bigr)
&= \frac{V(\rho) - v}{\tau},
\end{aligned}}
\end{equation}
where denote $\rho(x,t)$ as the vehicle density, $v(x,t)$ as the vehicle speed, $P'(\rho)=p'(\rho)/\rho$ as the pressure potential (with a typical choice $p(\rho)=\rho^\gamma,,\gamma>1$), $V(\rho)$ as the equilibrium fundamental–diagram speed, and $\tau>0$ as the relaxation time representing the behavioral time scale.
Without the source term ($\tau\to\infty$), equation \eqref{eq:arz} reduces to the
classical ARZ momentum balance
$\partial_t(v+P(\rho)) + v\,\partial_x(v+P(\rho))=0$.
The source accelerates drivers toward the equilibrium curve $v=V(\rho)$.

\subsubsection{Characteristic Variables}
ARZ model is a set of hyperbolic PDE that describe the process of density $\rho$ and velocity $v$ propagating along different characteristic lines on a road. To diagonalize this coupled system and explicitly utilize the properties of invariants along the characteristic line in the subsequent ARZ linearization and kernel function construction, we introduce the Riemann variable, also called Lagrangian markers \citep{aw2000resurrection, zhang2002non}:
\begin{align}
w_{1} &= v + P(\rho), \\
w_{2} &= v.
\end{align}
These two variables are precisely the combined coordinates that maintain constants along the two characteristic lines.

Under the new coordinates, the eigenvalues of the Jacobian matrix of the ARZ system flux correspond to the propagation speeds along the two characteristic lines \citep{zhang2002non}
\begin{align}
\lambda_{1}(\rho,v) &= v + \rho\,P'(\rho), \label{eq:characspeed1} \\
\lambda_{2}(\rho,v) &= v, \label{eq:characspeed2}
\end{align}
Consequently, along the characteristic curves \(dx/dt = \lambda_{i}(\rho,v)\), the corresponding Riemann variable \(w_{i}\) is (up to relaxation) constant in the inviscid limit.  Equivalently, in the absence of a source term:
\begin{equation}
\partial_{t}w_{i} + \lambda_{i}(\rho,v)\,\partial_{x}w_{i} = \frac{V(\rho) - v}{\tau},
\quad i=1,2.   
\end{equation}

These observations form the basis for the subsequent linearization and kernel construction.

\subsubsection{Perturbations around a Uniform Equilibrium}
In the ARZ, both the flux Jacobian matrix and the relaxation term depend on the current state $(\rho,v)$. If an operator is directly applied to it, it is not a linear operator and it is difficult to write a closed form of the GP covariance kernel function. In order to obtain the linear differential operator of constant coefficient, we adopt the linearization around the equilibrium \citep{belletti2015prediction}.

Specifically, we choose an equilibrium state $(\rho_0,v_0)$ with
\(
v_0 = V(\rho_0),
\)
and define small perturbations:
\begin{equation}
\delta w_i = w_i - w_{i,0}\,
\end{equation}
\begin{equation}
\delta \rho = \rho - \rho_0\,
\end{equation}
\begin{equation}
\delta v = v - v_0\,
\end{equation}

Since we are concerned with the small perturbations near the equilibrium state, these disturbances themselves are very small. Therefore, by introducing Taylor expansion and discarding the higher order \(\mathcal{O}(\delta^{2})\) terms, the description error of the first-order behavior is very small, and it is completely sufficient to characterize physical quantities such as the propagation speed. A first–order Taylor expansion of $P(\rho)$ about $\rho_0$ gives
\(
P(\rho) \approx P(\rho_0) + P'(\rho_0)\,\delta\rho,
\)
hence:
\begin{align}
\delta w_1 &= \delta v + P'(\rho_0)\,\delta\rho\,,\\
\delta w_2 &= \delta v\,.
\end{align}
Accordingly,
\begin{equation}
\delta\rho = \frac{\delta w_1-\delta w_2}{P'(\rho_0)}\,.
\end{equation}

\subsubsection{Linearized Relaxation Source Term}
We have already introduced the Riemann variable. Next, we will linearize the source term and perform the same processing on \(\tfrac{V(\rho)-v}{\tau}\) to maintain the form of constant coefficients.
Expanding around the equilibrium state \((\rho_{0},v_{0})\) and discard all \(\mathcal{O}\bigl((\delta\rho)^{2},\,(\delta v)^{2},\,\delta\rho\,\delta v\bigr)\) high terms, we have:
\begin{equation}
V(\rho) - v
  = \bigl[V(\rho_{0}) + V'(\rho_{0})\,\delta\rho + \mathcal{O}(\delta^{2})\bigr]
    - \bigl[v_{0} + \delta v\bigr]
  = V'(\rho_{0})\,\delta\rho \;-\;\delta v \;+\;\mathcal{O}(\delta^{2}).   
\end{equation}

Since \(\delta w_{1} = \delta v + P'(\rho_{0})\,\delta\rho\) and \(\delta w_{2} = \delta v\), we have:
\begin{equation}
\delta\rho = \frac{\delta w_{1} - \delta w_{2}}{P'(\rho_{0})},    
\end{equation}
Afterwards, we define:
\begin{equation}
\alpha := \frac{V'(\rho_{0})}{P'(\rho_{0})}
\end{equation}
\begin{equation}
\beta := 1 + \alpha
\end{equation}
After dropping \(\mathcal{O}(\delta^{2})\) terms, the relaxation source becomes as follows:
\begin{equation}
\frac{V(\rho) - v}{\tau}
  \;\approx\; 
  \frac{1}{\tau}\bigl[V'(\rho_{0})\,\delta\rho - \delta v\bigr]
  = \frac{1}{\tau}\bigl[\alpha\,(\delta w_{1} - \delta w_{2}) - \delta w_{2}\bigr]
  = \frac{1}{\tau}\bigl(\alpha\,\delta w_{1} - \beta\,\delta w_{2}\bigr).    
\end{equation}

Thus we get the linearized form:
\begin{equation}\label{eq:rhs-lin}
{%
\frac{V(\rho) - v}{\tau}
  \;\approx\;
  \frac{1}{\tau}\bigl(\alpha\,\delta w_{1} - \beta\,\delta w_{2}\bigr).}
\end{equation}

\subsubsection{Constant–Coefficient Linear System}
Building on the first-order linearization of the relaxation source term derived in the previous section, we proceed to incorporate the convection term, which has been decomposed at the equilibrium point with its characteristic velocity frozen. Along with the relaxation source term expressed in terms of constant coefficients, these components yield a linear PDE system characterized by constant coefficients.
The perturbation vector of the linear operator can be written as:
\begin{equation}
\bm W :=
\begin{pmatrix}
\delta w_1\\
\delta w_2
\end{pmatrix}    
\end{equation}
We then combine these two parts and write a convection-relaxation linear system with constant coefficients.
Below, we explain in detail why omitting this step leads to mixing of constant and perturbation terms, and how retaining only \(\delta w_{i}\) on the left–hand side ensures a correct first‐order system.
\paragraph{1. Decomposing \(w_{i}\) into equilibrium plus perturbation}  
By definition, for \(i=1,2\), 
\begin{equation}
w_{i}(x,t) \;=\; w_{i,0} + \delta w_{i}(x,t),    
\end{equation}

where
$
w_{1,0} = v_{0} + P(\rho_{0})$ and $w_{2,0} = v_{0}$ are constants (the values at equilibrium).  Consequently, we have:
\begin{equation}
\partial_{t}w_{i} 
= \partial_{t}(w_{i,0} + \delta w_{i}) 
= \underbrace{\partial_{t}w_{i,0}}_{=0} + \partial_{t}\,\delta w_{i} 
= \partial_{t}\,\delta w_{i},    
\end{equation}
\begin{equation}
\partial_{x}w_{i} 
= \partial_{x}(w_{i,0} + \delta w_{i}) 
= \underbrace{\partial_{x}w_{i,0}}_{=0} + \partial_{x}\,\delta w_{i} 
= \partial_{x}\,\delta w_{i}.    
\end{equation}

To guarantee that all constant terms drop out with only first-order perturbations, one way to write the advection term is:
\begin{equation}
\partial_{t}w_{i} + \lambda_{i}^{0}\,\partial_{x}w_{i},     
\end{equation}

Without first substituting \(w_{i}=w_{i,0}+\delta w_{i}\), then \(\partial_{t}w_{i}\) and \(\partial_{x}w_{i}\) would still contain the (constant) terms \(\partial_{t}w_{i,0}\) and \(\partial_{x}w_{i,0}\).  Since \(w_{i,0}\) is constant, \(\partial_{t}w_{i,0}=\partial_{x}w_{i,0}=0\),  one must verify this explicitly.  Otherwise, the residual \(\partial_{t}w_{i,0}\) or \(\partial_{x}w_{i,0}\) would remain in the linearized equation. Consequently, we have: 
\begin{align}
w_{i} &= w_{i,0} + \delta w_{i}, \\
\partial_{t} w_{i} &= \partial_{t}\,\delta w_{i}, \\
\partial_{x} w_{i} &= \partial_{x}\,\delta w_{i}.
\end{align}

\paragraph{2. Taylor–expanding and freezing characteristic speeds}  
Based on the equilibrium relations and the characteristic decomposition,  we have:
\begin{equation}
\lambda_{1}(\rho,v) 
= (v_{0} + \delta v) + (\rho_{0} + \delta\rho)\,P'\bigl(\rho_{0} + \delta\rho\bigr).    
\end{equation}

Expanding \(P'(\rho_{0} + \delta\rho)\) to first order, we have:
\begin{equation}
P'(\rho_{0} + \delta\rho) 
= P'(\rho_{0}) + P''(\rho_{0})\,\delta\rho + \mathcal{O}(\delta^{2}).    
\end{equation}

Hence,
\begin{equation}
\begin{aligned}
\lambda_{1}(\rho,v) 
&= v_{0} + \delta v 
  + (\rho_{0} + \delta\rho)\,\bigl[P'(\rho_{0}) + P''(\rho_{0})\,\delta\rho + \mathcal{O}(\delta^{2})\bigr]\\
&= v_{0} + \rho_{0}\,P'(\rho_{0})
  + \delta v + \rho_{0}\,P''(\rho_{0})\,\delta\rho + \delta\rho\,P'(\rho_{0})
  + \mathcal{O}(\delta^{2}).
\end{aligned}
\end{equation}

When forming a first–order system, we discard \(\mathcal{O}(\delta^{2})\).  Moreover, we fix the characteristic speed by setting:
\begin{align}
\lambda_{1}^{0} &= v_{0} + \rho_{0}\,P'(\rho_{0}) \\
\lambda_{2}^{0} &= v_{0}
\end{align}

and drop all remaining \(\delta\)-dependent corrections to \(\lambda_{1}\). Thus, in the linearized PDE, \(\lambda_{i}\) appears only as the constants \(\lambda_{i}^{0}\). \(\displaystyle \lambda_{2}^{0} = v_{0}\) is the pure convective or “vehicle” wave–speed.  It represents information that simply travels downstream at the local traffic speed \(v_{0}\). In addition, the \(\displaystyle \lambda_{1}^{0} = v_{0} + \rho_{0}\,P'(\rho_{0})\) is the pressure or congestion wave–speed.  Here \(P'(\rho_{0})\) quantifies how rapidly a small density perturbation propagates through the traffic.  Thus \(\lambda_{1}^{0}\) describes how a slowdown or “jam wave” moves relative to the vehicles.  If \(\rho_{0}\,P'(\rho_{0})>0\), congestion signals propagate faster (or upstream) relative to the flow, reflecting drivers’ reaction to density changes.

\paragraph{3. Assembling the linear advection term}  
Combining the two points above, the advection term for \(\delta w_{1}\) becomes:
\begin{equation}
\begin{aligned}
\partial_t w_1 + \lambda_1(\rho,v)\,\partial_x w_1
&= \partial_t \delta w_1 + \lambda_1^{0}\,\partial_x \delta w_1
\;+\;
\big(\partial_t w_{1,0} + \lambda_1^{0}\,\partial_x w_{1,0}\big)
\\
&\quad+\;
\Big[\,(\partial_\rho \lambda_1)_0\,\delta\rho + (\partial_v \lambda_1)_0\,\delta v\,\Big]\;\partial_x w_{1,0}
\;+\; \mathcal{O}(\delta^2).
\end{aligned}
\end{equation}
Since we linearize about a spatially uniform equilibrium,
$\partial_t w_{1,0}=0$ and $\partial_x w_{1,0}=0$ (with $v_0=V(\rho_0)$),
and we keep terms up to $\mathcal O(\delta)$,
we have $[\lambda_1^0+\mathcal O(\delta)]\,\partial_x\delta w_1
=\lambda_1^0\,\partial_x\delta w_1+\mathcal O(\delta^2)$,
so the equation reduces to
\begin{equation}
  \partial_t\delta w_1+\lambda_1^0\,\partial_x\delta w_1+\mathcal O(\delta^2).  
\end{equation}

\paragraph{4. Ensuring a strictly first‐order system}  
Based on the aforementioned derivation, we can arrive at the purely first-order linearized form with no residual constant or higher‐order pieces:
\begin{align}
w_{1}(\rho,v)        &= w_{1,0} + \delta w_{1} \\
\lambda_{1}(\rho,v) &= \lambda_{1}^{0} + \mathcal{O}(\delta)
\end{align}
The same argument applies to the \(\delta w_{2}\) equation.  
 
In summary, all \(\delta w_{i}\) appear explicitly, ensuring we retain only first‐order perturbations. Both the characteristic speeds and the source term have been Taylor‐expanded at \((\rho_{0},v_{0})\) and any \(\mathcal{O}(\delta^{2})\) discarded.
Equation~\eqref{eq:linPDE} is the final constant‐coefficient linear system.
\begin{equation}\label{eq:linPDE}
\begin{cases}
\partial_t\delta w_1 + \lambda_1^0\,\partial_x \delta w_1
= \dfrac{1}{\tau}\bigl(\alpha\,\delta w_1 - \beta\,\delta w_2\bigr),\\[6pt]
\partial_t\delta w_2 + \lambda_2^0\,\partial_x \delta w_2
= \dfrac{1}{\tau}\bigl(\alpha\,\delta w_1 - \beta\,\delta w_2\bigr).
\end{cases}
\end{equation}

\paragraph{5. Assembly of the Convection–Relaxation Operator} 
To simplify the form of the Equation~\eqref{eq:linPDE} and facilitate its subsequent embedding into the kernel function, we further define:
\begin{align}
\bm W   &:= \begin{pmatrix}\delta w_1\\\delta w_2\end{pmatrix}\,,\\
\Lambda &:= \mathrm{diag}(\lambda_1^0,\lambda_2^0)\,,\\
C       &:= \begin{pmatrix}
               \alpha & -\beta\\
               \alpha & -\beta
             \end{pmatrix}\,.
\end{align}

Then Equation~\eqref{eq:linPDE} can be written as:
\begin{equation}\label{eq:linPDE2}
\bigl[\partial_t I_2 + \Lambda\,\partial_x - \tfrac{1}{\tau} C \bigr]\bm W = \bm 0.
\end{equation}
Specifically, in our linearized operator, we have:  
\begin{equation}
    \mathcal{L} \;=\;\partial_{t} I \;+\;\Lambda\,\partial_{x} \;-\;\frac{1}{\tau}C,
\end{equation}
where the diagonal matrix is: 
\begin{equation}
    \Lambda 
= 
\begin{pmatrix}
  \lambda_{1}^{0} & 0 \\[4pt]
  0 & \lambda_{2}^{0}
\end{pmatrix}
\end{equation}

Equation~\eqref{eq:linPDE2} encodes two physically distinct wave‐speeds in the ARZ model. In \(\mathcal{L}\), the term \(\lambda_{1}^{0}\,\partial_{x}\) combined with the source‐matrix coefficients \(\tfrac{1}{\tau}(\alpha\,\delta w_{1} - \beta\,\delta w_{2})\) ensures that density‐driven disturbances \(\delta w_{1}\) propagate exactly at this jam‐wave speed.

Equation~\eqref{eq:linPDE} remains strictly hyperbolic because
$\lambda_1^0\neq\lambda_2^0$ unless $P'(\rho_0)=0$. The source couples the two characteristics but does not affect hyperbolicity.

\subsection{ARZ–Embedded Covariance Kernel}
\subsubsection{Base Kernel}
We adopt the stationary squared–exponential (SE) kernel:
\begin{align}
k_0\bigl((x,t),(x',t')\bigr)
  &= \sigma^2
     \exp\!\Biggl(
       -\tfrac{(x-x')^2}{2\ell_x^2}
       -\tfrac{(t-t')^2}{2\ell_t^2}
     \Biggr), \\
s  &= (x,t), \\
s' &= (x',t').
\end{align}
We select the SE kernel mainly because it is infinitely differentiable in both space and time dimensions, it can naturally capture the smooth variation characteristics of traffic density and velocity fields. Next, we apply the differential operator and its adjoint operator to the kernel function, and form the $2\times2$ block kernel:
\begin{equation}
K_0(s,s') := I_2 \otimes k_0(s,s')
          = \begin{pmatrix}
              k_0(s,s') & 0 \\ 0 & k_0(s,s')
            \end{pmatrix}.
\end{equation}

\subsubsection{Operator-Embedded Kernel}
In kernel method theory, positive-definite (PD) kernels maintain closure under the application of linear differential operators to each variable \citep{sarkka2011linear}. 
This property enables us to construct new valid kernels from fundamental kernel functions. Based on Equation ~\eqref{eq:linPDE}, we can write the differential operator as:
\begin{equation}
L_s
 := \partial_t\,I_2 + \Lambda\,\partial_x - \frac1\tau C
 = \begin{pmatrix}
     \partial_t+\lambda_1^0\partial_x-\tfrac{\alpha}{\tau}
     & \tfrac{\beta}{\tau} \\[6pt]
     -\tfrac{\alpha}{\tau}
     & \partial_t+\lambda_2^0\partial_x+\tfrac{\beta}{\tau}
   \end{pmatrix}.
\end{equation}
Its $L^2$–adjoint acting on the second argument is:
\begin{equation}
L_{s'}^{\!\top}
 = \partial_{t'}\,I_2 + \Lambda\,\partial_{x'}
   - \frac1\tau C^{\!\top}
 = \begin{pmatrix}
     \partial_{t'}+\lambda_1^0\partial_{x'}-\tfrac{\alpha}{\tau}
     & -\tfrac{\alpha}{\tau}\\[6pt]
     \tfrac{\beta}{\tau}
     & \partial_{t'}+\lambda_2^0\partial_{x'}+\tfrac{\beta}{\tau}
   \end{pmatrix}.
\end{equation}

Afterwards, the operator–embedded kernel is obtained:
\begin{equation}\label{eq:Klin}
{%
K_{\mathrm{lin}}(s,s')
   := L_s\bigl[K_0(s,s')\bigr]L_{s'}^{\!\top}}
\end{equation}
where $L_s$ and $L_{s'}^{\!\top}$ are linear differential operators acting on the first and second variables, respectively.

Equation~\eqref{eq:Klin} embeds the PDE constraints softly as any sample drawn from the GP prior satisfies $L_s\bm W(s)\equiv\bm 0$ exactly.

Further expanding the block form
$K_{\mathrm{lin}}=[k_{ij}]_{i,j=1}^{2}$, we have:
\begin{align}
k_{11}(s,s') &=
  K_{tt'} + (\lambda_1^0)^2 K_{xx'} + \lambda_1^0\!\left(K_{tx'}+K_{xt'}\right)
 - a\!\left(K_t + K_{t'} + \lambda_1^0 (K_x+K_{x'})\right)
 + a^2 K, \tag{41} \\[4pt]
k_{12}(s,s') &=
  K_{tt'} + \lambda_1^0\lambda_2^0 K_{xx'}
 + \lambda_1^0 K_{xt'} + \lambda_2^0 K_{tx'}
 - a\!\left(K_{t'} + \lambda_2^0 K_{x'}\right)
 + b\!\left(K_t + \lambda_1^0 K_x\right)
 - ab\,K, \tag{42} \\[4pt]
k_{21}(s,s') &=
  K_{tt'} + \lambda_1^0\lambda_2^0 K_{xx'}
 + \lambda_1^0 K_{tx'} + \lambda_2^0 K_{xt'}
 + b\!\left(K_{t'} + \lambda_1^0 K_{x'}\right)
 - a\!\left(K_t + \lambda_2^0 K_x\right)
 - ab\,K, \tag{43} \\[4pt]
k_{22}(s,s') &=
  K_{tt'} + (\lambda_2^0)^2 K_{xx'} + \lambda_2^0\!\left(K_{tx'}+K_{xt'}\right)
 + b\!\left(K_t + K_{t'} + \lambda_2^0 (K_x+K_{x'})\right)
 + b^2 K. \tag{44}
\end{align}
Finally, we have:
\begin{equation}
K_{\mathrm{lin}}(s,s') 
= \begin{pmatrix}
    k_{11}(s,s') & k_{12}(s,s') \\[6pt]
    k_{21}(s,s') & k_{22}(s,s')
  \end{pmatrix}.
\end{equation}

\subsubsection{Residual Kernel}
To compensate for the high-order and nonlinear effects discarded during the earlier linearization and velocity freezing, and to improve both the flexibility and numerical stability of the model, we incorporate an additional small-scale SE residual kernel into the total covariance:
\begin{equation}
    K_{\mathrm{res}}(s,s')
  := B_{\mathrm{res}}\,
     \sigma_{\mathrm{res}}^2
     \exp\!\Bigl[-\frac{(x-x')^2}{2\tilde\ell_x^2}
                 -\frac{(t-t')^2}{2\tilde\ell_t^2}\Bigr]
\end{equation}
where \(B_{\mathrm{res}}\) is the scaling factor of the residual kernel. With the introduction of the residual kernel, we can:
\begin{itemize}
  \item {Compensate higher-order nonlinearities:} The linearization
    retains only first-order terms around the equilibrium. Any
    second- or higher-order contributions are modelled by \(K_{\mathrm{res}}\).
  \item {Accommodate measurement noise:} Real-world observations of
    \(\rho\) and \(v\) include sensor error and data‐processing artifacts.
    \(K_{\mathrm{res}}\) provides the flexibility to explain discrepancies
    between the linearized PDE and noisy data.
  \item {Ensure numerical stability:} If \(K_{\mathrm{res}}\equiv 0\),
    slight deviations from \(\mathcal{L}\bm W =0\) in Equation~\eqref{eq:linPDE2} can make the GP posterior covariance
    matrix ill-conditioned. A nonzero nugget mitigates this and avoids
    overconfident, unstable estimates.
\end{itemize}

In Gaussian process modeling, if \(k_1\) and \(k_2\) are valid positive‐definite kernels, then their sum is also a valid kernel \citep{rasmussen2003gaussian, scholkopf2002learning}:
\begin{equation}
k(x, x') \;=\; k_1(x, x') \;+\; k_2(x, x')   
\end{equation}
Therefore, the total prior covariance kernel is
\begin{equation}
    K_{\mathrm{tot}} \;:=\; K_{\mathrm{lin}} + K_{\mathrm{res}}.
\end{equation}

\subsection{Gaussian–Process Inference}

\subsubsection{Variational Inference via Sparse Multi‐Output GP}
Although the posterior distribution of a Gaussian process admits a closed-form expression, computing the inverse and determinant of the $N \times N$ covariance matrix becomes computationally expensive and numerically unstable when $N$ is large. To address this issue, we adopt the sparse variational approximation \citep{titsias2009variational}, which significantly reduces computational complexity. Specifically, it introduces $M \ll N$ inducing points $\{\,\bm z_m=(x_m,t_m)\}_{m=1}^M$. We stack the two outputs at each location so that the inducing variable is \(\bm u\in\mathbb{R}^{2M}\) and the latent vector of all outputs at data locations \(\{s_i\}_{i=1}^{N}\) is \(\bm f=[f(s_1),\dots,f(s_N)]^\top\). From the total kernel \(K_{\mathrm{tot}}\) used throughout, we assemble the standard covariance blocks:
\begin{equation}
\mathbf{K}_{uu}=\bigl[K_{\mathrm{tot}}(\bm z_m,\bm z_{m'})\bigr]_{m,m'=1}^{M}
\end{equation}
\begin{equation}
\mathbf{K}_{uf}=\bigl[K_{\mathrm{tot}}(\bm z_m,s_i)\bigr]_{m=1,\dots,M}^{\,i=1,\dots,N}
\end{equation}
\begin{equation}
\mathbf{K}_{fu}=\mathbf{K}_{uf}^{\top}
\end{equation}
\begin{equation}
\mathbf{K}_{ff}=\bigl[K_{\mathrm{tot}}(s_i,s_j)\bigr]_{i,j=1}^{N}
\end{equation}
Projecting the joint prior through the inducing subspace yields the Schur complement:
\begin{gather}
\mathbf{Q}_{ff}^{\perp}=\mathbf{K}_{ff}-\mathbf{K}_{fu}\,\mathbf{K}_{uu}^{-1}\,\mathbf{K}_{uf},
\end{gather}
which captures the residual prior covariance not explained by the Nyström approximation.

\paragraph{Variational Approximation and Moments.}
We posit a Gaussian variational distribution over inducing variables, defined as:
\begin{gather}
q(\bm u)=\mathcal N(\bm m,\mathbf S),\qquad \bm m\in\mathbb R^{2M},\ \mathbf S\in\mathbb R^{2M\times 2M}\ \text{positive definite}.
\end{gather}
Given the variational distribution \(q(\bm u)\), Combining it with the conditional GP mapping from \(\bm u\) to \(\bm f\) implied by the prior gives a Gaussian marginal \(q(\bm f)\) with moments:
\begin{equation}
\mathbb{E}_{q}[\bm f]=\mathbf{K}_{fu}\,\mathbf{K}_{uu}^{-1}\,\bm m
\end{equation}
\begin{equation}
\mathrm{Cov}_{q}[\bm f]=\mathbf{Q}_{ff}^{\perp}
+\mathbf{K}_{fu}\,\mathbf{K}_{uu}^{-1}\,(\mathbf S-\mathbf{K}_{uu})\,\mathbf{K}_{uu}^{-1}\,\mathbf{K}_{uf}
\end{equation}
These expressions are employed both for predictions at the training locations and within the training objective.

\paragraph{Observation model.}
Each datum is modeled as a noisy measurement of the corresponding latent:
\begin{gather}
p(y_i\mid f(s_i))=\mathcal N\!\bigl(y_i\mid f(s_i),\,\sigma_{y,i}^2\bigr),\qquad
\sigma_{y,i}^2\in\{\sigma_\rho^2,\sigma_v^2\},
\end{gather}
where \(\sigma_\rho^2\) applies to density observations and \(\sigma_v^2\) applies to speed observations, while the kernel structure \(K_{\mathrm{tot}}\) remains unchanged.

\paragraph{ELBO}
We learn all kernel hyperparameters of \(K_{\mathrm{tot}}\) together with the variational parameters \((\bm m,\mathbf S)\) by maximizing the evidence lower bound \citep{titsias2009variational}:
\begin{gather}
\mathcal L=\mathbb{E}_{q(\bm u)}\!\bigl[\log p(\mathbf y\mid \bm u)\bigr]\;-\;\mathrm{KL}\!\bigl[q(\bm u)\,\|\,p(\bm u)\bigr].
\end{gather}
Under the Gaussian assumptions, both terms admit closed forms. The expected log–likelihood decomposes over observations as:
\begin{gather}
\mathbb{E}_{q(\bm u)}\!\bigl[\log p(\mathbf y\mid \bm u)\bigr]
=-\tfrac12\sum_{i=1}^N\!\Bigl[
\log\!\bigl(2\pi\,\sigma_{y,i}^2\bigr)
+\tfrac{(y_i-\mu_{f,i})^2+\Sigma_{f,ii}}{\sigma_{y,i}^2}
\Bigr],
\end{gather}
with \(\mu_{f,i}=[\,\mathbf{K}_{fu}\mathbf{K}_{uu}^{-1}\bm m\,]_i\) and \(\Sigma_{f,ii}=[\,\mathrm{Cov}_{q}(\bm f)\,]_{ii}\). The Kullback–Leibler divergence between \(q(\bm u)\) and the GP prior \(p(\bm u)=\mathcal N(\bm 0,\mathbf K_{uu})\) is:
\begin{gather}
\mathrm{KL}\!\bigl[q(\bm u)\,\|\,p(\bm u)\bigr]
=\tfrac12\Bigl[
\log\tfrac{\lvert \mathbf{K}_{uu}\rvert}{\lvert \mathbf S\rvert}
-2M+\mathrm{tr}\!\bigl(\mathbf{K}_{uu}^{-1}\mathbf S\bigr)
+\bm m^{\!\top}\mathbf{K}_{uu}^{-1}\bm m
\Bigr].
\end{gather}

\subsubsection{Optimization and Complexity}
We jointly optimize
$\bm m, \mathbf{S}$, the inducing inputs $\{\bm z_m\}_{m=1}^M$, and all kernel hyperparameters 
$\{\ell_x, \ell_t, \sigma,\\ \tilde\ell_x, \tilde\ell_t,$
$\sigma_{\mathrm{res}}, \sigma_\rho, \sigma_v, \tau, \alpha, \beta\}$
by maximizing $\mathcal{L}$.  Each ELBO evaluation requires computing \(\mathbf{K}_{uu}^{-1}\) (cost \(\mathcal{O}\bigl((2M)^{3}\bigr)\)) and forming \(\mathbf{K}_{fu}\) (cost \(\mathcal{O}(2N \cdot 2M)\)).  Because \(M\ll N\), the overall complexity per iteration is \(\mathcal{O}(N M^{2} + M^{3})\), which is much lower than the \(\mathcal{O}\bigl((2N)^{3}\bigr)\) needed for a full GP.
\subsubsection{Predictive Distribution}
Once the variational parameters are optimized, the predictive distribution at a new location \(s_{*}\) follows:
\begin{align}
q\bigl(f(s_*)\bigr) 
  &= \int p\bigl(f(s_*)\mid \bm u\bigr)\,q(\bm u)\,d\bm u
     = \mathcal{N}\bigl(\mu_*,\,\Sigma_*\bigr) \\
\mu_* 
  &= \mathbf{k}_{*u}\,\mathbf{K}_{uu}^{-1}\,\bm m \\
\Sigma_* 
  &= K_{\mathrm{tot}}(s_*,s_*) 
     - \mathbf{k}_{*u}\,\mathbf{K}_{uu}^{-1}\,\bigl(\mathbf{K}_{uu} - \mathbf{S}\bigr)\,
        \mathbf{K}_{uu}^{-1}\,\mathbf{k}_{u*}
\end{align}
where \(\mathbf{k}_{*u} = \bigl[K_{\mathrm{tot}}(s_{*},\bm z_{m})\bigr]_{m=1}^M\). The final predictive distribution for the observed fields $(\rho_*,v_*)$ augments the posterior covariance with the corresponding observation noise:
\begin{equation}
p(y_* \mid \mathcal{D}) 
  \approx \mathcal{N}\!\bigl(\mu_*,\,\Sigma_* + \mathrm{diag}(\sigma_\rho^2,\sigma_v^2)\bigr).
\end{equation}

Lastly, to obtain predictions for $(\rho,v)$ from $(w_1,w_2)$, linearize at $\rho_0$ and use:
\begin{equation}
    \begin{pmatrix}\mu_\rho\\ \mu_v\end{pmatrix}
=\begin{pmatrix}\rho_0-\dfrac{P(\rho_0)}{P'(\rho_0)}\\[4pt]0\end{pmatrix}
+\begin{pmatrix}\dfrac{1}{P'(\rho_0)} & -\dfrac{1}{P'(\rho_0)}\\[4pt] 0 & 1\end{pmatrix}
\begin{pmatrix}\mu_{w_1}\\ \mu_{w_2}\end{pmatrix}
\end{equation}
and propagate the covariance as:
\begin{align}
\Sigma_{\rho,v}=A\,\Sigma_w\,A^\top
\end{align}
\begin{align}
    A=\begin{pmatrix}\dfrac{1}{P'(\rho_0)} & -\dfrac{1}{P'(\rho_0)}\\[4pt] 0 & 1\end{pmatrix}.
\end{align}
Where \(A\) is the Jacobian of the mapping \((w_1,w_2)\mapsto(\rho,v)\) evaluated at the linearization point \(\rho_0\); \(\Sigma_{\rho,v}\in\mathbb{R}^{2\times2}\) is the predictive covariance of \((\rho,v)\) at the test location; 
\(\Sigma_w\in\mathbb{R}^{2\times2}\) is the posterior covariance of the latent pair \((w_1,w_2)\) at the same location.
\section{Case Study}
\subsection{Datasets}
We use the NGSIM (Next Generation Simulation) \citep{simulation2007us} and HighD \citep{krajewski2018highd} vehicle trajectory dataset to validate the effectiveness of the proposed method. NGSIM dataset records detailed trajectory information of vehicles on American highways 101, with a sampling frequency of 10 Hz. We screen the original data and select the vehicle data for subsequent processing. HighD dataset records detailed microscopic trajectory information for vehicles on German highways, captured with a sampling frequency of 25 Hz. To reconstruct the macroscopic traffic state from the microscopic vehicle trajectory data, we divide the entire observation range into regular spatiotemporal grids. Each original data point is mapped to the corresponding grid cell according to its temporal and spatial coordinates.

\subsection{Baseline Model and Evaluation Metrics}

In this paper, our baseline models include two state-of-the-art GP-based methods: GP-Rotated \citep{wu2024traffic} and adaptive smoothing interpolation method ASM \citep{treiber2011reconstructing}. The first baseline method is the Anisotropic Gaussian Process model. The core innovation of this method lies in the design of an anisotropic covariance kernel function that reflects the physical characteristics of traffic speed. The key hyperparameters are initialized as follows: the signal variance $\sigma^2$ is set to 0.2, and the likelihood noise variance $\sigma_n^2$ is set to 0.3. The anisotropic lengthscales in the rotated coordinate system are set to 150.0 (spatial dimension) and 13.0 (temporal dimension) respectively.

Moreover, ASM bypasses complex traffic flow modeling and directly reconstructs the velocity field through a directional spatio-temporal filtering. The key parameters were set as follows: the spatial resolution $\Delta x$ was 10 meters and the temporal resolution $\Delta t$ was 5 seconds. The characteristic spatial smoothing distance, $\sigma$, was set to 200 meters, and the temporal smoothing time, $\tau$, was set to 10 seconds.

We use mean absolute error (MAE) and root mean square error (RMSE) to evaluate  the performance. Let the spatial and temporal indices be $i = 1,\dots,S$ and $j = 1,\dots,T$, respectively.  
Define the valid set as:
\begin{equation}
  \Omega := \{(i,j)\mid y_{ij}\ \text{is available}\}, \qquad
  N = |\Omega|.
\end{equation}

For each valid grid point $(i,j)\in\Omega$, we denote the ground‑truth value
by $y_{ij}$ and the reconstructed value by $\hat y_{ij}$.
MAE and RMSE over the whole field are then defined as:
\begin{align}
\mathrm{MAE}
  &= \frac{1}{N}\,
     \sum_{(i,j)\in\Omega} \bigl|\,\hat y_{ij}-y_{ij}\bigr|, \label{eq:mae}\\
\mathrm{RMSE}
  &= \sqrt{%
     \frac{1}{N}\,
     \sum_{(i,j)\in\Omega} \bigl(\hat y_{ij}-y_{ij}\bigr)^2 }. \label{eq:rmse}
\end{align}
where $N = S\,T$. When speed $v$ and density $\rho$ are
evaluated separately, one simply substitutes
$y_{ij}\!=\!v_{ij}$, $\hat y_{ij}\!=\!\hat v_{ij}$ for speed, or
$y_{ij}\!=\!\rho_{ij}$, $\hat y_{ij}\!=\!\hat\rho_{ij}$ for density in
Eqs.\,\eqref{eq:mae}–\eqref{eq:rmse}.

\subsection{Speed and Density Estimation under Different Penetration Rate}

In this section, the spatio-temporal trajectories generated by probe vehicles are taken as the observation experiment objects. The expressive and extrapolative capabilities of each method for traffic speed and density fields are evaluated under varying penetration rates, aiming to assess reconstruction performance in large-scale scenarios.
The experimental results in the HighD dataset are reported in Tables~\ref{tab:highd_combined}, and those in the NGSIM dataset are reported in Tables~\ref{tab:ngsim_combined}. Across both datasets, the errors decrease rapidly with the increase in penetration rate. On HighD, ASM is slightly better at the extreme low end around five percent, but as penetration rate rises to ten to twenty percent the second–order PEGP--ARZ leads for both targets. For instance, the speed MAE decreases from approximately $0.80$ to $0.63$ km/h, while the density MAE declines from about $3.10$ to $2.48$ veh/km, suggesting that stronger physical constraints are particularly beneficial when observational data remain limited but informative. Once penetration rate exceeds roughly thirty percent, the first–order PEGP--LWR becomes dominant on HighD, with speed MAE near $0.49$--$0.38$\,km/h and density MAE near $1.85$--$1.44$\,veh/km at thirty to fifty percent, while the rotated GP stays consistently ahead of ASM. On NGSIM the pattern is similar in trend but the crossover points are later, errors fall by about forty percent from three to ten percent of penetration rates, PEGP--LWR and the rotated GP already surpasses ASM, and beyond twenty percent the gains are small. At the higher penetration rates, around thirty to forty percent, PEGP--ARZ attains the lowest speed and density errors on NGSIM, with speed MAE around $2.64$--$2.53$\,m/s and density MAE around $3.37$--$3.20$\,veh/km, while PEGP--LWR remains a close second and the rotated GP continues to outperform ASM. Taken together, both datasets reveal a consistent relationship that links prior knowledge with observed density and speed. Stronger physical knowledge prevails in the scarce to moderate regime. Once the observational data becomes sufficient, either the fist order PEGP--LWR model or the second-order PEGP--ARZ model emerges as the most effective compared to ASM and GP--Rotated.

\begin{figure}[htbp]
  \centering
  \includegraphics[width=0.8\textwidth]{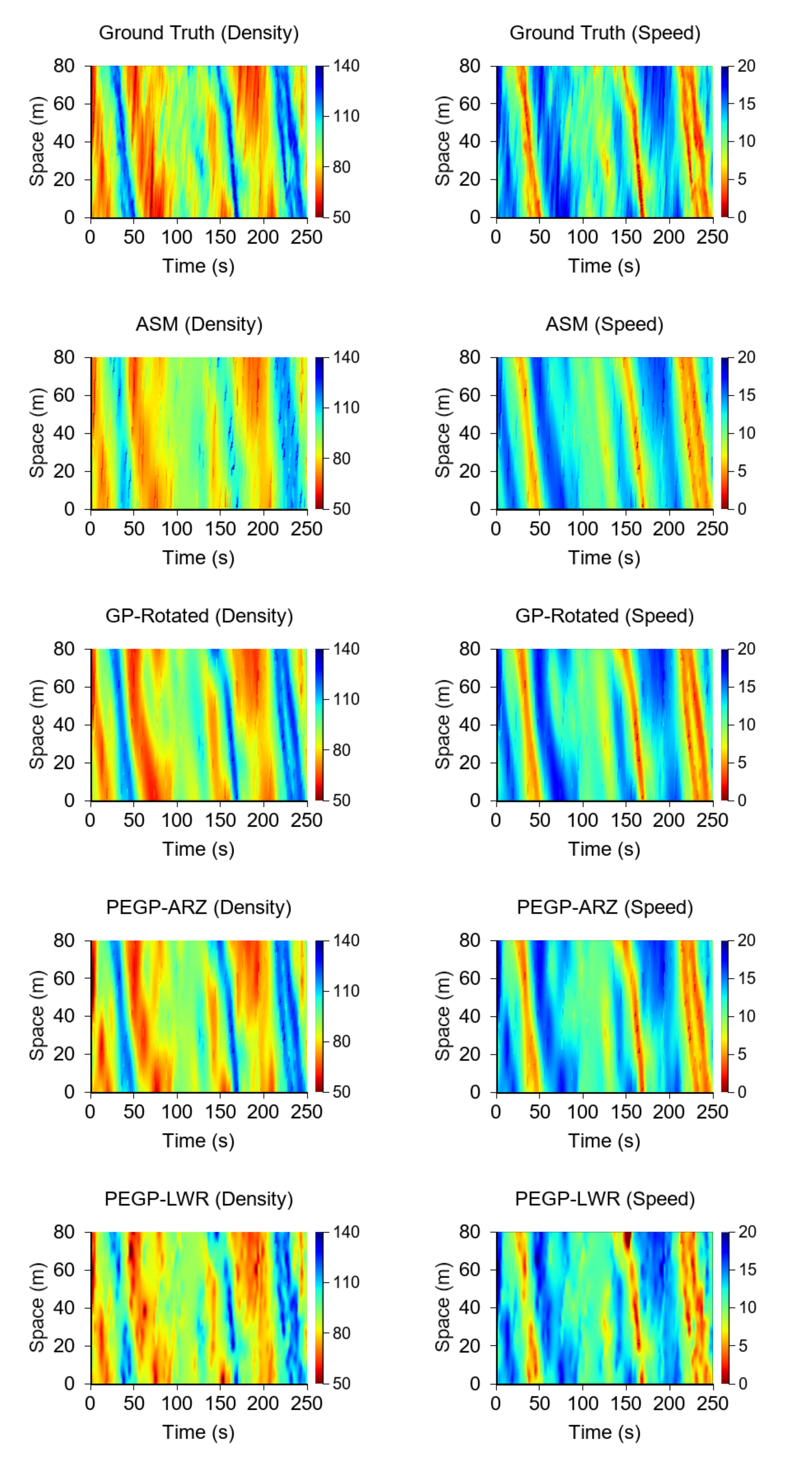}
  \caption{Speed and Density Reconstruction Under 10\% Penetration Rate (HighD)}
  \label{Trajdensityhighd}
\end{figure}

Figure~\ref{Trajdensityhighd} presents the Speed and Density Estimation under a 10\% penetration rate. The left panel illustrates the reconstructed density field, while the right panel depicts the corresponding velocity field. The four methods demonstrate distinct reconstruction behaviors on the HighD datasets. GP-Rotated captures the broad wave patterns but introduces local irregularities, which is consistent with its moderate error levels. ASM strengthens wave-like structures, yet the presence of noticeable artifacts leads to consistently higher errors compared with GP-based methods. In contrast, PEGP–ARZ enforces physical consistency and achieves the lowest errors at 10\% penetration rate, producing reconstructions that closely follow the wave fronts, though some fine details are lost. PEGP–LWR, derived from a first-order conservation law, provides greater flexibility in capturing local variations. At 10\% penetration rate, however, its overall reconstruction trend appears incomplete and the waves remain discontinuous. As the penetration rate increases, the reconstructions gradually improve and surpass PEGP--ARZ to become the most accurate method.

\begin{table}[H]
\footnotesize
\renewcommand{\arraystretch}{1.2}
\centering
\caption{Speed and Density Reconstruction Errors at Different Penetration Rates (HighD)}
\label{tab:highd_combined}
\begin{tabular}{clcccc}
\toprule
\multirow{2}{*}{Penetration Rate} & \multirow{2}{*}{Metric} & \multicolumn{4}{c}{Reconstruction Method} \\
\cmidrule(lr){3-6}
& & ASM & GP-Rotated & PEGP-LWR & PEGP-ARZ \\
\midrule
\multirow{2}{*}{5\%} & Speed & \textbf{1.07/1.40} & 1.09/1.52 & 1.28/1.47 & 1.09/1.51 \\
& Density & \textbf{3.95/5.25} & 4.02/5.60 & 4.57/6.70 & 4.20/5.75 \\
\addlinespace
\multirow{2}{*}{10\%} & Speed & 1.00/1.29 & 0.89/1.20 & 0.87/1.21 & \textbf{0.80/1.10} \\
& Density & 3.58/4.83 & 3.27/4.43 & 3.24/4.55 & \textbf{3.10/4.22} \\
\addlinespace
\multirow{2}{*}{20\%} & Speed & 0.92/1.22 & 0.69/0.93 & 0.69/1.13 & \textbf{0.63/0.90} \\
& Density & 3.12/4.38 & 2.56/3.46 & 2.50/4.04 & \textbf{2.48/3.42} \\
\addlinespace
\multirow{2}{*}{30\%} & Speed & 0.83/1.02 & 0.59/0.79 & \textbf{0.49/0.72} & 0.56/0.73 \\
& Density & 2.65/3.84 & 2.21/2.95 & \textbf{1.85/2.70} & 2.12/2.82 \\
\addlinespace
\multirow{2}{*}{40\%} & Speed & 0.72/0.84 & 0.50/0.65 & \textbf{0.42/0.60} & 0.50/0.67 \\
& Density & 2.13/3.25 & 1.81/2.42 & \textbf{1.56/2.25} & 1.91/2.53 \\
\addlinespace
\multirow{2}{*}{50\%} & Speed & 0.66/0.71 & 0.46/0.61 & \textbf{0.38/0.55} & 0.49/0.64 \\
& Density & 1.91/2.85 & 1.56/2.12 & \textbf{1.44/1.98} & 1.86/2.44 \\
\bottomrule
\multicolumn{6}{l}{\footnotesize Note: Values are presented as MAE/RMSE. Speed in km/h, Density in veh/km.}
\end{tabular}
\end{table}

\begin{table}[H]
\footnotesize
\renewcommand{\arraystretch}{1.2}
\centering
\caption{Speed and Density Reconstruction Errors at Different Penetration Rates (NGSIM)}
\label{tab:ngsim_combined}
\begin{tabular}{clcccc}
\toprule
\multirow{2}{*}{Penetration Rate} & \multirow{2}{*}{Metric} & \multicolumn{4}{c}{Reconstruction Method} \\
\cmidrule(lr){3-6}
& & ASM & Rotated & PEGP-LWR & PEGP-ARZ \\
\midrule
\multirow{2}{*}{3\%} & Speed &\textbf{ 5.34/9.02} & 5.69/8.80 & 5.66/8.79 & 5.80/9.01 \\
& Density & 6.96/10.23 & 7.07/10.85 & \textbf{6.92/10.81} & 7.25/11.28 \\
\addlinespace
\multirow{2}{*}{5\%} & Speed & 4.26/8.50 & 4.12/7.69 & 4.08/7.63 & \textbf{4.05/7.61} \\
& Density & 5.25/9.66 & 5.05/9.35 & \textbf{4.86/9.15} & 5.13/9.66 \\
\addlinespace
\multirow{2}{*}{10\%} & Speed & 3.69/7.88 & 3.25/6.82 & \textbf{3.22/6.43} & 3.24/6.67 \\
& Density & 4.35/8.98 & 4.14/8.72 & \textbf{4.01/8.42} & 4.07/8.35 \\
\addlinespace
\multirow{2}{*}{20\%} & Speed & 3.29/6.69 & 2.87/6.15 & 2.82/6.18 & \textbf{2.77/5.88} \\
& Density & 3.74/8.02 & 3.54/7.71 & \textbf{3.47/7.65} & 3.48/7.36 \\
\addlinespace
\multirow{2}{*}{30\%} & Speed & 3.04/6.32 & 2.76/6.04 & 2.71/5.98 & \textbf{2.64/5.92} \\
& Density & 3.59/7.89 & 3.48/7.76 & \textbf{3.42/7.49} & 3.37/7.52 \\
\addlinespace
\multirow{2}{*}{40\%} & Speed & 2.86/6.05 & 2.68/5.91 & 2.65/5.86 & \textbf{2.53/5.57} \\
& Density & 3.46/7.65 & 3.31/7.37 & 3.27/7.38 & \textbf{3.20/7.00} \\
\bottomrule
\multicolumn{6}{l}{\footnotesize Note: Values are presented as MAE/RMSE. Speed in m/s, Density in veh/km.}
\end{tabular}
\end{table}

\subsection{Speed and Density Estimation in a Four-Loop Detector Scenario}
After evaluating the different penetration rates of probe vehicles, we simulate the sparse cross-section data obtained by only a small number of loop detectors, and evaluate the model performance in order to demonstrate its potential for application in environments with limited sensor availability.
In detail, we set four loop detectors in fixed sections ($y=60,\,180,\,300,\,420$\,m) along the longitudinal $600\,\mathrm{m}$ section and used the velocity\slash density sequence of these sections as the observation. 

Table~\ref{tab:loop performance comparison} reports the velocity and density reconstruction errors. For the velocity field, GP-Rotated yields a MAE of 5.45,m/s, which is higher than those of the PEGP approaches. PEGP–LWR achieves the lowest velocity error (MAE=4.57,m/s), indicating that, based on loop-detector observation, the first-order conservation model provides the most effective physical prior for velocity estimation. The density reconstruction results show that GP-Rotated yields a relatively high RMSE of 10.87,veh/km. PEGP–ARZ improves performance by achieving lower errors, with density MAE of 5.90,veh/km and RMSE of 9.01,veh/km. Among four methods, PEGP–LWR achieves the lowest density MAE (5.01,veh/km) and RMSE (7.93,veh/km), indicating that with only four loop-detector observations, a simple first-order model can still provide sufficient physical priors for density estimation.

\begin{figure}[htbp]
        \centering
        \includegraphics[width=\textwidth]{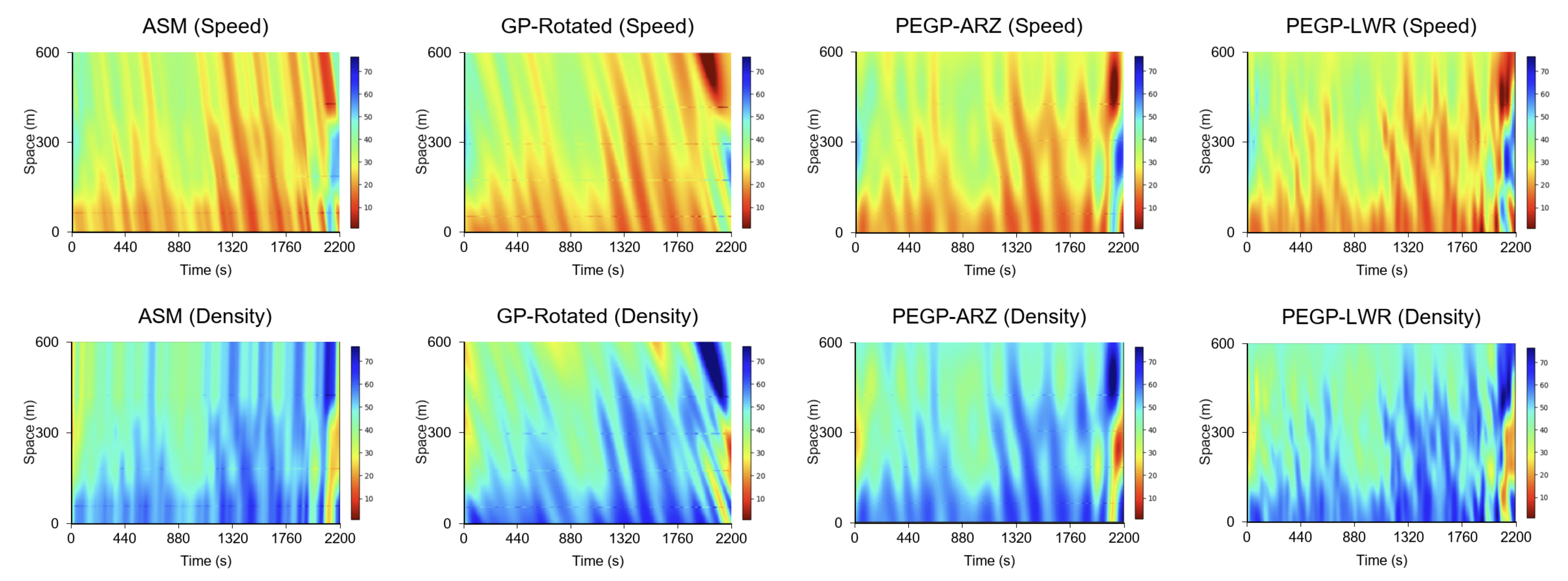}
        \caption{Speed and Density Estimation in a Four-Loop Detector Scenario}
        \label{loopdetector}
\end{figure}

\begin{table}[H]
\footnotesize
\renewcommand{\arraystretch}{0.8}
\centering
\caption{Reconstruction Errors for Four Loop Detectors Comparison of Different Methods}
\label{tab:loop performance comparison}
\begin{tabular}{lcccc}
\toprule
\multirow{2}{*}{Reconstruction Method} & \multicolumn{2}{c}{Speed} & \multicolumn{2}{c}{Density} \\
\cmidrule(lr){2-3} \cmidrule(lr){4-5}
& MAE (m/s) & RMSE (m/s) & MAE (veh/m) & RMSE (veh/m) \\
\midrule
ASM & 4.82 & 8.38 & 6.16 & 9.31 \\
GP-Rotated & 5.45 & 8.45 & 6.70 & 10.87 \\
PEGP-ARZ  & 4.66 & 7.11 & 5.90 & 9.01 \\
PEGP-LWR & \textbf{4.57} & \textbf{7.06} & \textbf{5.01} & \textbf{7.93} \\
\bottomrule
\end{tabular}
\end{table}
Figure~\ref{loopdetector} presents the reconstruction results from four methods with four loop-detector observations, where the first row shows velocity reconstruction and the second row shows density reconstruction. The four methods exhibit distinct characteristics in reconstructing the velocity field. GP-Rotated reproduces the global trend but introduces spurious patterns in the 2000–2200s segment, corresponding to a MAE of 5.45,m/s. ASM strengthens wave patterns but at the cost of noticeable artifacts, whereas PEGP–ARZ emphasizes physical consistency but sacrifices some fine details. PEGP-LWR achieves the most balanced reconstruction, maintaining smoothness and detail simultaneously. The density reconstructions in Figure~\ref{loopdetector} show consistent patterns across methods. ASM captures the overall trend but underestimates the fluctuation magnitudes, GP-Rotated introduces local irregularities, PEGP–ARZ maintains physical plausibility with moderate clarity, and PEGP–LWR provides the most balanced performance in terms of accuracy and smoothness.

Overall, the four-loop detector experiment highlights the importance of incorporating physical priors, as evidenced by the lower errors of PEGP–ARZ and PEGP–LWR compared with the non-PEGP baselines. To further illustrate, PEGP–ARZ improves speed estimation by leveraging the physical constraints imposed by its relaxation term, which regulates the magnitude of fluctuations. For PEGP–LWR, the simplified first-order conservation law aligns more closely with this scenario, leading to the lowest errors. Moreover, PEGP--LWR performs better than PEGP--ARZ, consistent with the trend on HighD at higher penetration rates, which may reflect a form of physical overfitting. Strong priors contribute to stable velocity estimation but risk oversmoothing the state field when overly restrictive, whereas weak priors or highly flexible kernels provide greater adaptability.

\subsection{Ablation study}
From the kernel definitions introduced in the methodology, PEGP--ARZ and PEGP--LWR can be viewed as complementary components within an additive residual kernel. This decomposition allows us to analyze the distinct contribution of each part. Although experiments on the NGSIM dataset show alternating advantages between PEGP--ARZ and PEGP--LWR, the HighD dataset consistently exhibits the pattern that PEGP--ARZ dominates at low penetration while PEGP--LWR takes over at high penetration rate. This raises the question of why these regime differences arise, which we address by probing two complementary aspects of the posterior:
(i) Energy attribution: how much of the final prediction is explained by the physics component versus the learned residual;
(ii) representation overlap: how
similar or orthogonal the physics and residual subspaces are, independent of the final prediction direction.
This section of the study is based on the HighD dataset.
\subsubsection{Physics/residual Share Analysis }
A direct way to analyze the contribution of each kernel component is to quantify how much of the final prediction originates from the physics term versus the residual term.
We decompose the posterior mean $\mu=\mu_{\text{phys}}+\mu_{\text{res}}$ at sampled points and report:
(i) aligned shares $S_{\text{phys}}=\mu^\top\mu_{\text{phys}}/\|\mu\|^2$,
$S_{\text{res}}=\mu^\top\mu_{\text{res}}/\|\mu\|^2$,
and (ii) energy shares $E_{\text{phys}}=\|\mu_{\text{phys}}\|^2/\|\mu\|^2$,
$E_{\text{res}}=\|\mu_{\text{res}}\|^2/\|\mu\|^2$. Here, the aligned shares characterize the directional alignment of physics versus residuals with the final prediction, whereas the energy shares quantify their relative magnitudes. The detailed definitions and derivations are provided in~\ref{energy Share analysis}.
The share results in Table~\ref{tab:components} highlight a clear contrast: PEGP--ARZ predictions are dominated by the physics component, while PEGP--LWR relies heavily on residuals at low penetration rates and gradually balances the two. For PEGP--ARZ, the aligned physics share for $u=w_2$ rises from $72.5\%$ at $p=5\%$ to $98.6\%$ at $p=20\%$, with energy shares above $80\%$ and joint ratios consistently below $0.5$. This indicates that both direction and magnitude of the prediction are almost constrained by the underlying physics. By contrast, PEGP--LWR at $p=5\%$ shows only $9.5\%$ aligned physics for speed and $10.6\%$ for density, with residual energies above $80\%$ and a joint ratio of $9.3$, revealing that early predictions are driven by weakly aligned residuals. As penetration rate increases to $30$--$50\%$, PEGP--LWR physics shares rise to about $40\%$, residual energy falls to $ 42\%$, and the joint ratio decreases toward $1.4$, showing that residuals become well aligned with the target.

These trends align closely with the TSE results on HighD datesets. At low penetration rates ($p=5$--$10\%$), PEGP--ARZ achieves the lowest velocity errors (MAE $1.09$ at $5\%$, $0.80$ at $10\%$) and competitive density errors, consistent with its high physics alignment that stabilizes predictions under sparse observation. PEGP--LWR, dominated by misaligned residuals in this regime, yields larger errors (e.g.\ velocity MAE $1.28$ at $5\%$). At higher penetration ($p \ge 20\%$), as PEGP--LWR’s residuals become aligned, it surpasses PEGP--ARZ in both variables: velocity MAE decreases to $0.49$ at $30\%$ and $0.38$ at $50\%$, compared with PEGP--ARZ $0.56$ and $0.49$, while density errors also become lower for PEGP--LWR ($1.44$ at $50\%$ versus PEGP--ARZ $1.86$). To sum up, the share analysis explains why PEGP--ARZ is robust at low coverage, but PEGP--LWR benefits more from additional data and achieves superior accuracy at medium to high penetration rates.

\begin{table}[H]
\footnotesize
\renewcommand{\arraystretch}{0.8}
\centering
\caption{Component shares vs penetration $p$. All shares in \%; ratios are res:phys.}
\label{tab:components}
\scriptsize
\begin{tabular}{@{}r
cccc cccc c@{}}
\toprule
 & \multicolumn{4}{c}{\textbf{PEGP--LWR $\rho$}} & \multicolumn{4}{c}{\textbf{PEGP--LWR $u$}} & \multicolumn{1}{c}{\textbf{PEGP--LWR}} \\
\cmidrule(lr){2-5}\cmidrule(lr){6-9}\cmidrule(lr){10-10}
$p$ (\%) & phys & res & $E_{\rm phys}$ & $E_{\rm res}$ & phys & res & $E_{\rm phys}$ & $E_{\rm res}$ & joint ratio \\
\midrule
 5  & 10.6 & 89.4 &  1.9 & 80.7 &  9.5 & 90.5 &  1.5 & 82.5 & 9.31 \\
10  & 22.4 & 77.6 &  9.0 & 64.2 & 19.0 & 81.0 &  6.7 & 68.8 & 3.69 \\
20  & 40.9 & 59.1 & 25.1 & 43.3 & 38.3 & 61.7 & 22.3 & 45.7 & 1.57 \\
30  & 39.6 & 60.4 & 24.6 & 45.5 & 41.5 & 58.5 & 26.3 & 43.3 & 1.43 \\
40 & 41.2 & 58.8 & 25.8 & 43.3 & 42.7 & 57.3 & 27.7 & 42.4 & 1.54 \\
50  & 41.2 & 58.8 & 25.8 & 43.3 & 42.7 & 57.3 & 27.7 & 42.4 & 1.36 \\
\bottomrule
\end{tabular}

\vspace{0.5em}

\begin{tabular}{@{}r
cccc cccc c@{}}
\toprule
 & \multicolumn{4}{c}{\textbf{PEGP--ARZ $w_1$ (= u+P($\rho$))}} & \multicolumn{4}{c}{\textbf{PEGP--ARZ $w_2$ (= u)}} & \multicolumn{1}{c}{\textbf{PEGP--ARZ}} \\
\cmidrule(lr){2-5}\cmidrule(lr){6-9}\cmidrule(lr){10-10}
$p$ (\%) & phys & res & $E_{\rm phys}$ & $E_{\rm res}$ & phys & res & $E_{\rm phys}$ & $E_{\rm res}$ & joint ratio \\
\midrule
 5  & 60.1 & 39.9 & 39.6 & 19.5 & 72.5 & 27.5 & 56.2 & 11.1 & 0.51 \\
10  & 69.4 & 30.6 & 52.1 & 13.2 & 77.7 & 22.3 & 63.9 &  8.5 & 0.35 \\
20  & 84.1 & 15.9 & 72.1 &  4.0 & 89.6 & 10.4 & 81.5 &  2.3 & 0.15 \\
30  & 94.3 &  5.7 & 89.1 &  0.5 & 98.6 &  1.4 & 97.4 &  0.1 & 0.03 \\
40  & 89.3 & 10.7 & 80.5 &  2.0 & 93.2 &  6.8 & 87.6 &  1.2 & 0.09 \\
50  & 87.0 & 13.0 & 76.8 &  2.8 & 91.0 &  9.0 & 83.9 &  1.8 & 0.12 \\
\bottomrule
\end{tabular}

\vspace{0.25em}
\footnotesize Note: All raw numbers are retained. PEGP--LWR reports density/speed and joint; PEGP--ARZ reports $w_1(=u+P(\rho))$, $w_2(=u)$ and joint. The joint ratio columns show the res:phys ratio.
\end{table}

\subsubsection{Physics/residual Similarity Analysis }
Building on the physics/residual mean decomposition underlying the share metrics, we further examine the similarity between the subspaces spanned by physics and residual contributions, irrespective of magnitude. For each penetration rates $p$ and regime (FREE or CONGESTED), we draw $m$ spatio–temporal samples $\{(x_j, t_j)\}_{j=1}^m$ , evaluate the physics- and residual-induced posterior means at these points, de-standardize them to physical units, and center across samples to obtain paired feature matrices $X$ and $Y$. 
In each reported row, we present one pair, corresponding to per-task measures for PEGP--LWR and per-output measures for PEGP--ARZ where PEGP--PEGP--ARZ is expressed in terms of $(w_1,w_2)$. Conceptually, these diagnostics provide a distinct perspective from the share analysis by focusing on the geometry of physics and residual subspaces rather than their attribution magnitudes. Whereas the share metrics quantify how much of the final prediction is attributed to physics versus residual components, the Centered Kernel Alignment (CKA) \citep{kornblith2019similarity} and principal angles \citep{bjorck1973numerical} capture the intrinsic similarity or complementarity between the physics- and residual-induced subspaces, independent of their overall magnitudes. 
With linear kernels, the CKA between $X$ and $Y$ is
\begin{equation}
    \mathrm{CKA}(X,Y)=
\frac{\lVert X_c^\top Y_c\rVert_F^2}{\lVert X_c^\top X_c\rVert_F\,\lVert Y_c^\top Y_c\rVert_F},
\end{equation}
where $X_c$ and $Y_c$ denote column-centered features, 
$\mathrm{CKA}\in[0,1]$ is invariant to global rescaling and orthogonal rotations and increases with subspace similarity. To complement this scalar index, we compute principal angles by first orthonormalizing $X_c$ and $Y_c$ via QR to obtain orthonormal bases $Q_X, Q_Y$, and then performing a Singular Value Decomposition (SVD) of 
$Q_X^\top Q_Y$: 
\begin{equation}
   Q_X^\top Q_Y = U\,\mathrm{diag}(\sigma_1,\ldots,\sigma_k)\,V^\top ,
\end{equation}
from which the principal angles are obtained as: 
\begin{equation}
   \theta_i = \arccos(\sigma_i)\ \ \text{(in degrees)} .
\end{equation}
In the reported results, we summarize the smallest and largest among 
the top five principal angles, denoted as ``Min$^\circ$'' and ``Max$^\circ$''.
Sampling respects FREE/CONGESTED masks prior to selecting the $n$ points, and all entries use a single draw with $n=400$ to match the table specification.

Table~\ref{tab:combined-cka} shows that PEGP--PEGP--ARZ exhibits substantially higher similarity between physics and residual-induced subspaces than PEGP--LWR. In the free region, the CKA of PEGP--PEGP--ARZ decreases from $0.651$ at $p=5\%$ to $0.458$ at $p=50\%$, while the principal angles remain small, 
Min$^\circ$ increasing from $3.64$ to $8.76$ and Max$^\circ$ from $7.26$ to $14.88$. In the congested region, the CKA of PEGP--PEGP--ARZ decreases from $0.655$ at $p=5\%$ to $0.474$ at $p=50\%$, while the principal angles remain small, Min$^\circ$ increasing from $5.06$ to $9.58$ and Max$^\circ$ from $6.84$ to $15.88$. PEGP--LWR, by contrast, exhibits markedly lower CKA values and larger principal angles. 
In the free region, CKA remains around $0.27$–$0.32$, with Min$^\circ$ increasing from $11.54$ to $20.75$ and Max$^\circ$ from $22.78$ to $30.81$. 
In the congested region, CKA is $0.286$–$0.340$, with angles generally larger, 
Min$^\circ$ ranging from $14.01$ to $20.84$ and Max$^\circ$ from $29.60$ to $35.06$. Building upon the result, PEGP--ARZ’s high CKA and small angles align with its large aligned- and energy-physics shares and tiny joint ratios: physics and residual point in similar directions, so the total mean is physics-dominated across $p$, explaining the strong robustness at low coverage and the comparatively modest error reductions with increasing $p$. Second, PEGP--LWR exhibits lower CKA values and larger principal angles, indicating that the residuals span directions complementary to the physics subspace, particularly under congested conditions. As $p$ increases, these complementary residuals contribute more directly to the final prediction, reflected in the declining joint ratio reported in the shares table. However, they remain dissimilar to the physics subspace, as evidenced by the lack of increase in CKA. Taken together, this alignment explains the sharper MAE and RMSE improvements of PEGP--LWR at medium to high coverage.
In short, PEGP--PEGP--ARZ couples residuals that are nearly colinear with physics, reinforcing its physics-dominated bias. In contrast, PEGP--LWR learns residuals that are more orthogonal to the physics subspace and, once sufficiently supervised, can transform this orthogonality into a complementary signal that reduces errors, particularly in congested regimes.
\begin{table}[H]
\centering
\footnotesize
\caption{CKA$(\text{phys},\text{res})$ and principal angles by model and region.}
\label{tab:combined-cka}
\scriptsize
\begin{minipage}{0.48\textwidth}
\centering
\begin{tabular}{@{}r ccc|ccc@{}}
\toprule
 & \multicolumn{3}{c}{\textbf{FREE Region}} & \multicolumn{3}{c}{\textbf{CONGESTED Region}} \\
\cmidrule(lr){2-4}\cmidrule(lr){5-7}
$p$ (\%) & CKA & Min$^\circ$ & Max$^\circ$ & CKA & Min$^\circ$ & Max$^\circ$ \\
\midrule
\multicolumn{7}{l}{\textbf{PEGP-LWR (per task)}} \\
\midrule
 5  & 0.3187 & 11.54 & 22.78 & 0.3404 & 14.01 & 30.09 \\
10  & 0.3137 & 15.03 & 25.79 & 0.3373 & 15.82 & 32.03 \\
20  & 0.2819 & 21.05 & 32.09 & 0.3009 & 19.61 & 33.02 \\
30  & 0.2678 & 21.98 & 32.88 & 0.2858 & 21.20 & 35.06 \\
40  & 0.2776 & 20.39 & 31.44 & 0.2959 & 19.42 & 29.60 \\
50  & 0.2772 & 20.75 & 30.81 & 0.2957 & 20.84 & 32.05 \\
\bottomrule
\end{tabular}
\end{minipage}%
\hfill
\begin{minipage}{0.48\textwidth}
\centering
\begin{tabular}{@{}r ccc|ccc@{}}
\toprule
 & \multicolumn{3}{c}{\textbf{FREE Region}} & \multicolumn{3}{c}{\textbf{CONGESTED Region}} \\
\cmidrule(lr){2-4}\cmidrule(lr){5-7}
$p$ (\%) & CKA & Min$^\circ$ & Max$^\circ$ & CKA & Min$^\circ$ & Max$^\circ$ \\
\midrule
\multicolumn{7}{l}{\textbf{PEGP-PEGP--ARZ (per output)}} \\
\midrule
 5  & 0.6514 &  3.64 &  7.26 & 0.6553 &  5.06 &  6.84 \\
10  & 0.5510 &  6.69 & 11.09 & 0.5626 &  7.06 & 11.52 \\
20  & 0.5115 &  7.10 & 12.68 & 0.5257 &  7.70 & 14.11 \\
30  & 0.4930 &  7.91 & 13.49 & 0.5070 &  8.69 & 13.87 \\
40  & 0.4504 &  8.96 & 15.21 & 0.4662 & 10.08 & 18.67 \\
50  & 0.4585 &  8.76 & 14.88 & 0.4737 &  9.58 & 15.88 \\
\bottomrule
\end{tabular}
\end{minipage}

\vspace{0.5em}
\footnotesize \parbox{\textwidth}{Note: Min$^\circ$ and Max$^\circ$ refer to the smallest and largest (of top 5) principal angles in degrees between physical and residual subspaces. Higher CKA values indicate greater subspace alignment.}
\end{table}

\subsection{Uncertainty Quantification}
\begin{figure}[htbp]
  \centering
    \centering
    \includegraphics[width=\textwidth]{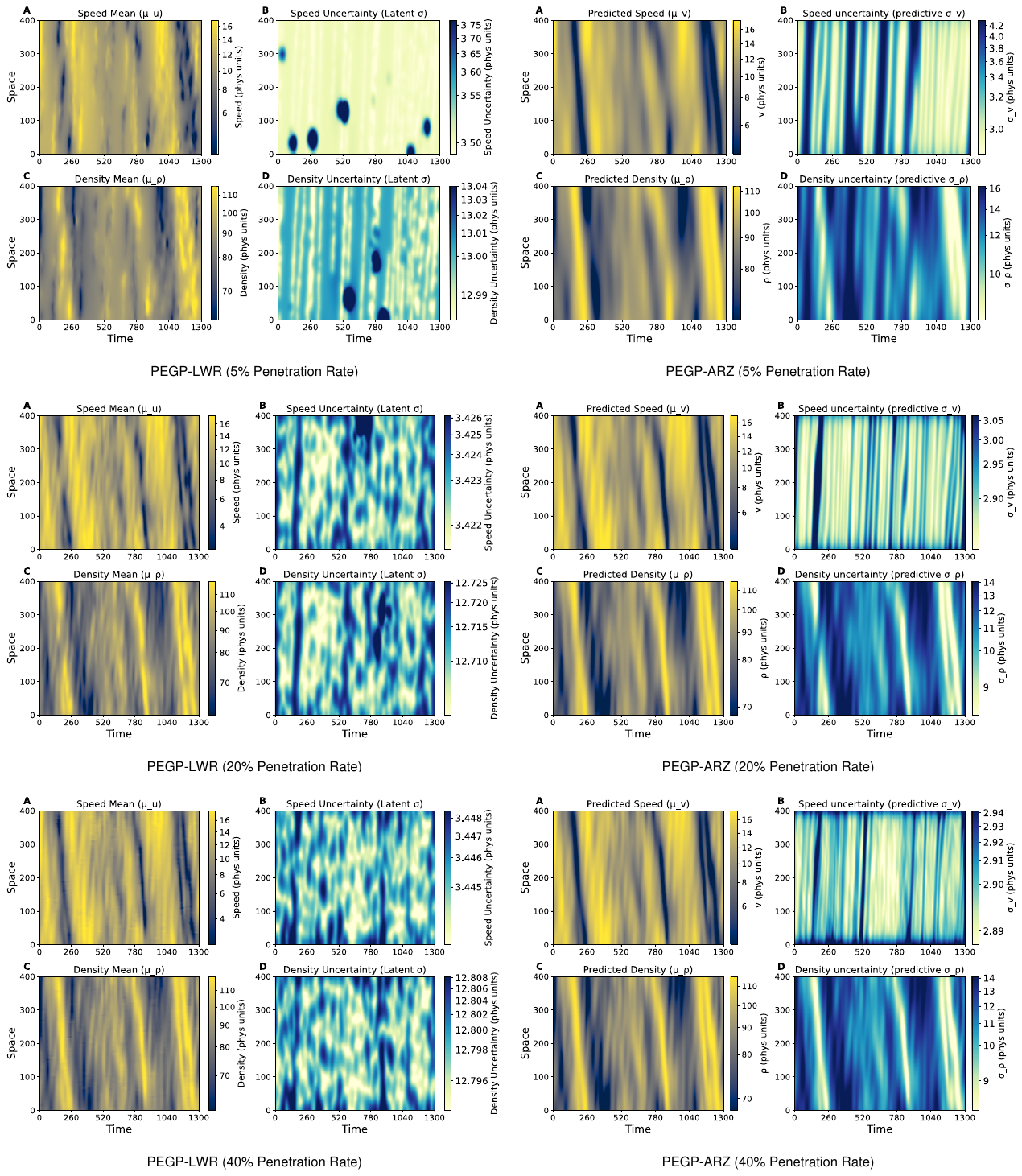}
    \caption{PEGP--ARZ Uncertainty Quantification Under several Penetration Rates (HighD)}
    \label{ARZ-UQ}
\end{figure}

The distinctive strength of Gaussian processes lies in their ability not only to estimate traffic states but also to provide a natural quantification of
uncertainty through the posterior variance. This endogenous estimation of
confidence levels represents a core advantage of the GP framework.
The detailed definitions and derivations of mean and uncertainty quantification are provided in~\ref{uncertainty quantification}. In Figure~\ref{ARZ-UQ}, the left column shows the predicted means and the right column corresponding uncertainty maps across penetration rates from 5\% to 40\%. Both PEGP–ARZ and PEGP–LWR recover the dominant spatio–temporal waves, with mean fields becoming smoother and sharper as the penetration rate increases. The two models exhibit 
fundamentally different structures. PEGP--ARZ generates variance bands that are aligned 
with the propagating waves and, although they gradually weaken, they remain visible even at higher penetration. In contrast, PEGP--LWR reduces to an almost constant field with only slight local variation in sparsely observed regions.

This sharp visual contrast reflects fundamental differences in how the two models propagate uncertainty. PEGP–ARZ maps uncertainty from the Riemann invariants into the physical variables, whereas PEGP–LWR constructs uncertainty directly in $(\rho,u)$. As a result, their predictive covariances are shaped by distinct mechanisms, which explains the amplified, wave-aligned structures in PEGP–ARZ and the nearly constant uncertainty fields in PEGP–LWR. The reason is that PEGP--ARZ is trained on the invariants $w_1,w_2$ and then mapped to the physical variables $\rho,u$ through a nonlinear pressure law $P(\rho)=\tfrac12\rho^2$. Additionally, the predictive covariance in $(\rho,u)$ is well approximated by the Jacobian pushforward:
\begin{align}
\sigma_{v,\mathrm{arz}}^{2}(t,x) &
= j_u\,\Sigma^{\text{tot}}_{w}(t,x)\,j_u^\top
= \bigl(\Sigma^{\text{tot}}_{w}(t,x)\bigr)_{22},
\\[4pt]
\sigma_{\rho,\mathrm{arz}}^{2}(t,x) 
&\approx j_\rho\,\Sigma^{\text{tot}}_{w}(t,x)\,j_\rho^\top
= 
\begin{bmatrix} \tfrac{1}{\rho} & -\tfrac{1}{\rho} \end{bmatrix}
\Sigma^{\text{tot}}_{w}(t,x)
\begin{bmatrix} \tfrac{1}{\rho} \\[2pt] -\tfrac{1}{\rho} \end{bmatrix}.
\end{align}
where $j_u=[\,0,\,1\,]$ is the Jacobian row for speed; 
$j_\rho=[\,1/\rho,\,-1/\rho\,]$ is the Jacobian row for density;
$\Sigma^{\text{tot}}_{w}(t,x)$ is the
prediction–space covariance obtained by adding the likelihood noise to the posterior predictive covariance; 
$\bigl(\Sigma^{\text{tot}}_{w}(t,x)\bigr)_{22}$ denotes the $(2,2)$ entry of $\Sigma^{\text{tot}}_{w}$.
Detailed definitions and proofs can be found in the ~\ref{uncertainty quantification} from~\eqref{lwrUQ1} to ~\eqref{lwrUQ2}.

Hence, the different forms of $j_u$ and $j_\rho$ directly explain the patterns seen in Figure~\ref{ARZ-UQ}. 
For speed, $j_u$ is constant, so the variance $\sigma_v^2(t,x)$ is simply $\bigl(\Sigma^{\text{tot}}_{w}(t,x)\bigr)_{22}$. This means the uncertainty in $u$ does not get reshaped by the Jacobian and remains fairly uniform across space, matching the nearly constant fields in the plots. 
In contrast, the density variance depends on $j_\rho$, which changes with the predictive mean density. 
Where $\rho$ is small or varies rapidly, the Jacobian acts like a magnifier that amplifies uncertainty by mixing contributions from both $w_1$ and $w_2$. 
As a result, $\sigma_\rho^2(t,x)$ shows strongly wave-aligned propagation, exactly as seen in the visualizations.

However, PEGP--LWR models $\rho$ and $u$ directly with smooth kernels, which suppresses directional amplification and yields much nearly constant uncertainty fields. With denser observations, the latent variances are strongly reduced, so that the predictive variances approach a constant baseline. Specifically, for the speed and density of predictive variance we obtain:
\begin{equation}
\sigma_{v,\text{lwr}}^{2}(t,x)=s_u^{2}\,\bigl(\sigma^{2\,(u)}_z(t,x)+\lambda_{u,z}\bigr).
\end{equation}
\begin{equation}
\sigma_{\rho,\text{lwr}}^{2}(t,x)=s_r^{2}\,\bigl(\sigma^{2\,(r)}_z(t,x)+\lambda_{r,z}\bigr).
\end{equation}
where $s_r,s_u$ are the de--standardization scales for density and speed; 
$\sigma^{2\,(r)}_z(t,x)$ and $\sigma^{2\,(u)}_z(t,x)$ are the latent posterior predictive variances in standardized space at $(t,x)$; 
$\lambda_{r,z},\lambda_{u,z}$ are the standardized likelihood noise variances.
The precise definitions and their derivations are given in the ~\ref{uncertainty quantification} from~\eqref{arzUQ1} to ~\eqref{arzUQ2}.

Accordingly, when smooth kernels are used and the penetration rate increases, 
the latent variances $\sigma^{2\,(r)}_z(t,x)$ and $\sigma^{2\,(u)}_z(t,x)$ 
become very small compared to the noise levels. 
This relationship can be expressed as:
\[
\sigma^{2\,(r)}_z(t,x)\ll\lambda_{r,z},\qquad 
\sigma^{2\,(u)}_z(t,x)\ll\lambda_{u,z}.
\]
After de--standardization, the predictive variances are therefore dominated by the noise terms,
leading to the approximation:
\[
\sigma_{\rho,\text{lwr}}^{2}(t,x)\approx s_r^2\lambda_{r,z},\qquad 
\sigma_{v,\text{lwr}}^{2}(t,x)\approx s_u^2\lambda_{u,z}.
\]

As a result, the uncertainty fields stabilize around a nearly constant baseline. In the visualizations, 
this is expressed as a broadly uniform uncertainty in both $\rho$ and $u$, with only minor residual 
fluctuations coming from small latent terms. Unlike PEGP--ARZ, there is no Jacobian 
transformation to amplify uncertainty along wave fronts, so the variance maps remain 
largely homogeneous without directional amplification.
In summary, the two models perform similarly in mean reconstruction, but PEGP–ARZ delivers uncertainty estimates that are richer and more physically interpretable, which makes it the better choice when reliable uncertainty quantification and diagnostic value are required.

\section{Conclusion}
This study introduces a Physics-Embedded Gaussian process framework with ARZ and LWR operators, which consistently outperform non-physics baselines. The framework is built by first linearizing the ARZ and LWR traffic flow models and applying the obtained linear coefficient differential operator to the covariance kernel of the Gaussian process, thereby embedding the structural prior of the conservation-relaxation and flow-velocity relationship at the kernel level. We further construct a multi-output covariance matrix that captures cross-correlations between the speed and density, allowing the GP to jointly model density and velocity waves in a unified framework. To assess the effectiveness of this construction, we evaluate the framework on large-scale trajectory datasets, including HighD and NGSIM. The experiments reveal a clear dependence on penetration rate. PEGP–ARZ is a more reliable choice under sparse observation or when calibrated and interpretable uncertainty is required, whereas PEGP–LWR achieves lower errors once penetration rate becomes moderate to high. Specifically, on the HighD dataset, errors decrease rapidly as penetration rate increases, and PEGP--ARZ performs best around 10\% to 20\%  while PEGP--LWR becomes preferable at 30\% to 50\% percent. On the NGSIM dataset, the trend is similar but the crossover occurs later and is weaker, with speed errors remaining largely comparable and PEGP--LWR providing only a mild advantage on density as penetration rate grows. In the Four-Loop Detector experiment, PEGP–LWR and PEGP–ARZ both surpass the non-physics alternatives. This indicates that embedding the ARZ or LWR operators as priors introduces physical knowledge that improve reconstruction quality, particularly under sparse sensor conditions.

To explain these results, the ablation analysis of aligned and energy shares, joint ratios, CKA, and principal angles shows that PEGP--ARZ residuals are nearly colinear with the physics subspace, reflected in high physics shares, higher CKA values, and smaller principal angles. This stabilizes learning when data are scarce. In contrast, PEGP--LWR residuals remain more orthogonal to physics, reflected in lower CKA values and larger principal angles. With sufficient observation, the residuals align more closely with the final prediction than with the physics, thereby providing complementary information that supports consistent improvements as the penetration rate increases. The uncertainty analysis further highlights a qualitative difference in variance propagation. PEGP--ARZ propagates latent covariance from the Riemann invariants into the physical variables $\rho$ and $u$ via the Jacobian pushforward, yielding wave-aligned and time-oriented uncertainty structures with diagnostic value. In contrast, PEGP--LWR models $\rho$ and $u$ directly with smoother kernels, resulting in nearly constant variance fields.

Although the proposed operator-embedded kernel method in this study performed well in the TSE experiment of a single road section, it still has the following limitations. First, it has only been verified in the scenarios of a single traffic flow direction, and its adaptability to complex road networks has not yet been investigated. Second, when the boundary conditions of linear operators change, it may be because the high-order nonlinear terms of the model are ignored, resulting in underestimated errors. Finally, although the computational efficiency has been significantly improved through sparse variational inference and matrix chunked computation, the model training and inference are still relatively long compared to the baseline model, and are not friendly enough to the real-time TSE scenarios. Future work can be improved in these aspects: expanding the form of operators in multi-branch road networks, introducing boundary absorption or nonlinear residual compensation, and optimizing approximate inference algorithms to further reduce response delay.

\newpage

\bibliographystyle{plainnat}

\bibliography{cas-refs}    

\newpage
\appendix
\section{Energy Share Analysis}\label{energy Share analysis}
Let $X=\{(x,t)\}$ be query locations and $Z$ the inducing set. For a multi-output GP with inducing variables,

\begin{equation}
\boldsymbol{\mu}(X)=K_{XZ}\boldsymbol{\alpha}
\end{equation}

\begin{equation}
\boldsymbol{\alpha}=(K_{ZZ})^{-1}\boldsymbol{\mu}(Z)
\end{equation}

where $K_{XZ}$ and $K_{ZZ}$ are the task-structured kernels at $(X,Z)$ and $(Z,Z)$. We split the kernel additively into \emph{physics} and \emph{residual} blocks,

\begin{equation}
K_{XZ}=K^{\text{phys}}_{XZ}+K^{\text{res}}_{XZ}
\end{equation}

\begin{equation}
\boldsymbol{\mu}_{\text{phys}}=K^{\text{phys}}_{XZ}\boldsymbol{\alpha}
\end{equation}

\begin{equation}
\boldsymbol{\mu}_{\text{res}}=K^{\text{res}}_{XZ}\boldsymbol{\alpha}
\end{equation}

so that $\boldsymbol{\mu}=\boldsymbol{\mu}_{\text{phys}}+\boldsymbol{\mu}_{\text{res}}$ holds elementwise.

PEGP--LWR: Outputs are $(\rho,u)$. The physics kernel applies bidirectional LWR operators to a shared base kernel $k_0$:

\begin{equation}
K^{\text{phys}}_{XZ}=w_f\,L_f L_f' k_0+(1-w_f)\,L_b L_b' k_0,\quad w_f\in[0,1]
\end{equation}

A linear task coupling with vector $[\,1,\ v'\,]$ induces a block structure proportional to
$\begin{psmallmatrix}1&v'\\ v'&{v'}^2\end{psmallmatrix}$; residual terms are task-diagonal SEs on $(\rho,\rho)$ and $(u,u)$ blocks. For either target (density or speed), the corresponding rows of $\boldsymbol{\mu}_{\text{phys}}$ and $\boldsymbol{\mu}_{\text{res}}$ give the physics and residual contributions.

PEGP--ARZ: Latent outputs are $w_1=u+P(\rho)$ and $w_2=u$. The physics kernel applies linearized ARZ operators (with cross-blocks) to $k_0$:

\begin{equation}
K^{\text{phys}}_{XZ}=\begin{bmatrix}L_1\\ L_2\end{bmatrix}\!\begin{bmatrix}L_1' & L_2'\end{bmatrix}k_0
\end{equation}

and residuals are task-diagonal SEs. Speed is read off as $u=w_2$; density is recovered via

\begin{equation}
\rho=P^{-1}(w_1-w_2)
\end{equation}

Thus $\mu_{\text{phys}/\text{res}}^{(u)}$ equals the $w_2$ component of $\boldsymbol{\mu}_{\text{phys}/\text{res}}$, and
$\mu_{\text{phys}/\text{res}}^{(\rho)}$ follows by applying $P^{-1}$ to the corresponding $(\mu^{(w_1)},\mu^{(w_2)})$ pair.

On a set of sampled space–time points (after per-task de-standardization), we stack the corresponding vectors of total, physical, and residual components as:
\begin{equation}
\boldsymbol{\mu},\ \boldsymbol{\mu}_{\text{phys}},\ \boldsymbol{\mu}_{\text{res}}\in\mathbb{R}^m
\end{equation}
\begin{equation}
\boldsymbol{\mu}=\boldsymbol{\mu}_{\text{phys}}+\boldsymbol{\mu}_{\text{res}}
\end{equation}
The aligned share of the physical component is defined as:
\begin{equation}
S_{\text{phys}}=\frac{\boldsymbol{\mu}^\top\boldsymbol{\mu}_{\text{phys}}}{\|\boldsymbol{\mu}\|_2^2}
\end{equation}
and similarly, that of the residual component as: 
\begin{equation}
S_{\text{res}}=\frac{\boldsymbol{\mu}^\top\boldsymbol{\mu}_{\text{res}}}{\|\boldsymbol{\mu}\|_2^2}
\end{equation}

By construction, the aligned shares satisfy:
\begin{align}
S_{\text{phys}}+S_{\text{res}}=1
\end{align}

For comparison, we also compute energy shares, which measure the normalized squared magnitudes of each component:
\begin{equation}
E_{\text{phys}}=\frac{\|\boldsymbol{\mu}_{\text{phys}}\|_2^2}{\|\boldsymbol{\mu}\|_2^2}
\end{equation}
\begin{equation}
E_{\text{res}}=\frac{\|\boldsymbol{\mu}_{\text{res}}\|_2^2}{\|\boldsymbol{\mu}\|_2^2}
\end{equation}
These quantities are related through:
\begin{equation}
\|\boldsymbol{\mu}\|_2^2=\|\boldsymbol{\mu}_{\text{phys}}\|_2^2+\|\boldsymbol{\mu}_{\text{res}}\|_2^2
+2\,\boldsymbol{\mu}_{\text{phys}}^\top\boldsymbol{\mu}_{\text{res}}
\end{equation}

Aligned shares can be negative if a component is anti-aligned with $\boldsymbol{\mu}$.

On the joint task we compute an aligned-share ratio: for PEGP--LWR, we stack $(\rho,u)$; for PEGP--ARZ, we stack $(w_1,w_2)$ (density is shown separately via $P^{-1}$). With stacked vectors $\tilde{\boldsymbol{\mu}}$, $\tilde{\boldsymbol{\mu}}_{\text{phys}}$, and $\tilde{\boldsymbol{\mu}}_{\text{res}}$,

\begin{equation}
\text{Joint ratio}=\frac{S_{\text{res}}}{S_{\text{phys}}}
=\frac{\tilde{\boldsymbol{\mu}}^\top\tilde{\boldsymbol{\mu}}_{\text{res}}}
       {\tilde{\boldsymbol{\mu}}^\top\tilde{\boldsymbol{\mu}}_{\text{phys}}}\,
\end{equation}

which summarizes the residual-to-physics directional contribution on the joint task.

\section{Uncertainty Quantification}\label{uncertainty quantification}
In the PEGP--LWR variant, the GP is defined directly on the outputs $(\rho,u)$, hence both means and uncertainties are read from the marginal predictive distribution and then de-standardized. 
In the PEGP--ARZ variant, the GP is defined on the Riemann invariants $(w_1,w_2)$ with a physically motivated mapping to $(\rho,u)$. The mean of $u$ comes directly from $w_2$, while the mean of $\rho$ is obtained by a nonlinear transformation of $(w_1-w_2)$. 
Predictive uncertainties are obtained from the predictive covariance of $(w_1,w_2)$ via delta-method propagation through the nonlinear map to $(\rho,u)$.

\subsection{PEGP--LWR: targets, standardization, and predictive moments (with noise).}
We train a two--output GP directly on $(\rho,u)$ with per--task standardization:
\begin{equation}\label{lwrUQ1}
y^{(r)}_z=\frac{\rho-m_r}{s_r}.
\end{equation}
\begin{equation}
y^{(u)}_z=\frac{u-m_u}{s_u}.
\end{equation}

At a location $(t,x)$, the GP yields standardized predictive means and variances 
$\mu^{(r)}_z, \mu^{(u)}_z$ and $\sigma^{2\,(r)}_z, \sigma^{2\,(u)}_z$. 
Transforming back to physical units gives the mean fields:
\begin{equation}
\mu_{\rho}(t,x)=m_r+s_r\,\mu^{(r)}_z(t,x).
\end{equation}
\begin{equation}
\mu_{v}(t,x)=m_u+s_u\,\mu^{(u)}_z(t,x).
\end{equation}

For latent (noise--free) uncertainty, variances follow by rescaling:
\begin{equation}
\sigma_{\rho,\text{lat}}^{2}(t,x)=s_r^{2}\,\sigma^{2\,(r)}_z(t,x).
\end{equation}
\begin{equation}
\sigma_{v,\text{lat}}^{2}(t,x)=s_u^{2}\,\sigma^{2\,(u)}_z(t,x).
\end{equation}

Including observation noise with standardized per--task noise variances 
$\lambda_{r,z},\lambda_{u,z}$ (i.e., $\Lambda_z=\mathrm{diag}(\lambda_{r,z},\lambda_{u,z})$), the observation--space predictive variances are:
\begin{equation}
\sigma_{\rho,\text{obs}}^{2}(t,x)=s_r^{2}\,\bigl(\sigma^{2\,(r)}_z(t,x)+\lambda_{r,z}\bigr).
\end{equation}
\begin{equation}\label{lwrUQ2}
\sigma_{v,\text{obs}}^{2}(t,x)=s_u^{2}\,\bigl(\sigma^{2\,(u)}_z(t,x)+\lambda_{u,z}\bigr).
\end{equation}

where $m_r,m_u$ are the training means of $(\rho,u)$; 
$s_r,s_u>0$ are the corresponding scale factors; 
$(\mu^{(\cdot)}_z,\sigma^{2\,(\cdot)}_z)$ are standardized posterior predictive moments; 
$\lambda_{r,z},\lambda_{u,z}$ are likelihood noise variances in the standardized space.

\subsection{PEGP--ARZ: invariants, standardization, and predictive moments (with noise).}
We train on the invariants $w_1=u+P(\rho)$ and $w_2=u$ with joint standardization. 
Let the standardized predictive Gaussian at $(t,x)$ be:
\begin{equation}\label{arzUQ1}
\bigl(\mu^{\text{pred}}_{w,z},\Sigma^{\text{pred}}_{w,z}\bigr).
\end{equation}
\begin{equation}
\mu^{\text{pred}}_{w,z}=\begin{bmatrix}\mu_{1,z}\\ \mu_{2,z}\end{bmatrix}.
\end{equation}
\begin{equation}
\Sigma^{\text{pred}}_{w,z}=
\begin{bmatrix}s_{11,z}&s_{12,z}\\ s_{12,z}&s_{22,z}\end{bmatrix}.
\end{equation}

With diagonal scale $D=\mathrm{diag}(s_1,s_2)$ and mean $m=(m_1,m_2)$, de--standardization gives:
\begin{equation}
\mu^{\text{pred}}_{w}=m+D\,\mu^{\text{pred}}_{w,z}.
\end{equation}
\begin{equation}
\Sigma^{\text{pred}}_{w}=D\,\Sigma^{\text{pred}}_{w,z}\,D.
\end{equation}

If the likelihood noise in standardized invariant space is $\Lambda_{w,z}$, then in physical units:
\begin{equation}
\Lambda_{w}=D\,\Lambda_{w,z}\,D.
\end{equation}
\begin{equation}
\Sigma^{\text{tot}}_{w}(t,x)=\Sigma^{\text{pred}}_{w}(t,x)+\Lambda_{w}.
\end{equation}

where $P(\cdot)$ is the pressure law; 
$m$ and $D$ are the per--component mean and scale used for standardization/de--standardization; 
$\Sigma^{\text{pred}}_{w}$ is the latent posterior predictive covariance in $(w_1,w_2)$; 
$\Lambda_{w}$ is the observation--noise covariance in the same space; 
$\Sigma^{\text{tot}}_{w}$ denotes latent plus noise covariance and will be used in the Jacobian pushforward below.
\paragraph{Mapping and Jacobian pushforward (with noise).}
Define the deterministic map from invariants to physical variables
\begin{align}
(\rho,u) &= g(w_1,w_2) = \bigl(f(w_1,w_2),\,w_2\bigr), \\
f(w_1,w_2) &= \sqrt{\,2\,\max(w_1-w_2,0)\,}.
\end{align}
At the predictive mean $\mu_w=(\mu_{w_1},\mu_{w_2})$, the Jacobian of $g$ is:
\begin{align}
J_g(\mu_w) &=
\begin{pmatrix}
\partial f/\partial w_1 & \partial f/\partial w_2\\[2pt]
0 & 1
\end{pmatrix}_{\mu_w}
=
\begin{pmatrix}
1/\rho & -1/\rho\\[4pt]
0 & 1
\end{pmatrix}, \\
\rho &= \sqrt{\,2(\mu_{w_1}-\mu_{w_2})\,}.
\end{align}
Let $\Sigma^{\text{pred}}_{w}(t,x)$ denote the latent (noise--free) predictive covariance of $(w_1,w_2)$ at $(t,x)$, 
and let $\Lambda_w$ be the observation--noise covariance in the $(w_1,w_2)$ space. 
Then the observation--space predictive covariance in $(\rho,u)$ is approximated by
\begin{equation}
\Sigma^{\text{obs}}_{\rho,u}(t,x)
\;\approx\;
J_g(\mu_w)\,\bigl(\Sigma^{\text{pred}}_{w}(t,x)+\Lambda_w\bigr)\,J_g(\mu_w)^\top.
\end{equation}
Where $g$ is the deterministic ARZ mapping; $J_g(\mu_w)$ is its Jacobian evaluated at the predictive mean $\mu_w$; 
$\Sigma^{\text{pred}}_{w}$ is the GP posterior predictive covariance for $(w_1,w_2)$; 
$\Lambda_w$ is the (possibly diagonal) likelihood noise covariance in the invariant space.If the GP is trained in standardized units, de--standardize before applying $J_g$: $\mu_w=m+D\,\mu_{w,z}$ and $\Sigma^{\text{pred}}_{w}=D\,\Sigma^{\text{pred}}_{w,z}\,D$, with $D=\mathrm{diag}(s_1,s_2)$.

\paragraph{predictive variances (with noise).}
Write $J_g(\mu_w)=\begin{psmallmatrix} j_\rho \\ j_u \end{psmallmatrix}$ with 
$j_\rho=[\,1/\rho,\,-1/\rho\,]$ and $j_u=[\,0,\,1\,]$, and let 
$\Sigma^{\text{tot}}_{w}(t,x)=\Sigma^{\text{pred}}_{w}(t,x)+\Lambda_w$. Then
\begin{align}\label{arzUQ2}
\sigma_v^{2}(t,x) &= \mathrm{Var}[u](t,x)
= j_u\,\Sigma^{\text{tot}}_{w}(t,x)\,j_u^\top
= \bigl(\Sigma^{\text{tot}}_{w}(t,x)\bigr)_{22},
\\[4pt]
\sigma_\rho^{2}(t,x) 
&\approx j_\rho\,\Sigma^{\text{tot}}_{w}(t,x)\,j_\rho^\top
= 
\begin{bmatrix} \tfrac{1}{\rho} & -\tfrac{1}{\rho} \end{bmatrix}
\Sigma^{\text{tot}}_{w}(t,x)
\begin{bmatrix} \tfrac{1}{\rho} \\[2pt] -\tfrac{1}{\rho} \end{bmatrix}.
\end{align}
where $\sigma_v^{2}$ and $\sigma_\rho^{2}$ denote the observation--space predictive variances for speed and density, respectively; 
$\Sigma^{\text{tot}}_{w}$ is the sum of latent predictive covariance and likelihood noise in the invariant space; 
$j_u$ and $j_\rho$ are the corresponding Jacobian rows of $J_g(\mu_w)$.

\section{Linearization of LWR Model and Construction of Covariance Kernel and PEGP--LWR Predictive Distribution}
\label{subsec:lwr_kernel_cn}
In this section, the overall process of the linearization framework, from uniform equilibrium perturbation and first-order Taylor expansion, freezing coefficients, defining linear operators and applying them to the base kernel to construct operator-Embedded kernels, and adding residual kernels to form total covariance, is completely consistent with the PEGP--ARZ method mentioned earlier. Therefore, a brief framework is provided here.
\subsection{LWR Conservation Equation}
The LWR model \cite{lighthill1955kinematic, richards1956shock} only contains the conservation of density $\rho(x,t)$:
\begin{align}\label{eq:lwr_original_cn}
\partial_t \rho + \partial_x q(\rho) &= 0\,,\\
q(\rho)                                &= \rho\,V(\rho)\,.
\end{align}

\subsection{Uniform Equilibrium State and Small perturbation}
To linearize the model, we first select a uniform equilibrium state characterized by a constant density $\rho_0>0$ and the corresponding equilibrium velocity is:
\begin{align}
   \ 
v_0:=V(\rho_0).
\
\end{align}
Next, we introduce small perturbations around this steady state to capture local deviations in density and velocity:
\begin{align}\label{eq:lwr_perturb_cn}
\rho &= \rho_0 + \delta\rho\,,\\
v    &= v_0 + \delta v = V(\rho_0) + V'(\rho_0)\,\delta\rho\,.
\end{align}
Substituting these expressions into the governing equations and retaining only the first-order terms in
 $\mathcal O(\delta\rho)$
 item yields the linearized system used in subsequent derivations.
\subsection{Taylor Expansion and Freezing Wave Velocity}
Perform the first-order expansion of the flux $q(\rho)=\rho V(\rho)$at $\rho_0$:
\begin{align}\label{eq:lwr_flux_taylor_cn}
q(\rho)      &\approx q(\rho_0) + q'(\rho_0)\,\delta\rho\,,\\
q'(\rho_0)   &= V(\rho_0) + \rho_0\,V'(\rho_0) =: \lambda_0\,.
\end{align}

Substitute Equation \eqref{eq:lwr_flux_taylor_cn}
\eqref{eq:lwr_original_cn} and discarding the higher-order terms, a constant coefficient linear PDE can be obtained:
\begin{equation}\label{eq:lwr_linear_pde_cn}
\partial_t\delta\rho + \lambda_0\,\partial_x\delta\rho \; = \; 0,
\end{equation}
Here, $\lambda_0$ represents the LWR characteristic wave velocity.

\subsection{Linear Operators}
To construct the physics-embedded covariance kernel, we first define the linear differential operator corresponding to the LWR model as:
\begin{equation}\label{eq:lwr_operator_cn}
\mathcal L_{LWR} \; : = \; \partial_t + \lambda_0\,\partial_x,
\quad\Longrightarrow\quad
\mathcal L_{LWR}\,\delta\rho = 0.
\end{equation}
Its $L^2$ adjoint operator is
$\mathcal L_{s'-LWR}^{\!\top}
 = \partial_{t'} + \lambda_0\,\partial_{x'}$.
To embed this operator within the Gaussian process prior, we begin with a basic scalar kernel that serves as the latent covariance.
Specifically, we select the squared exponential (SE) kernel of the form:
\begin{align}\label{eq:lwr_se_base_cn}
k_0(s,s') &= \sigma^2
\exp\! \Bigl[
-\tfrac{(x-x')^2}{2\ell_x^2}
-\tfrac{(t-t')^2}{2\ell_t^2}
\Bigr],\\
s &= (x,t).
\end{align}

\paragraph{Action Operator}
Since the positive definite kernel remains positive definite after applying linear differential operators to each independent variable separately \cite{sarkka2011linear}. it is defined:
\begin{equation}\label{eq:lwr_operator_kernel_cn}
K_{\mathrm{lin-LWR}}(s,s')
:= \mathcal L_{s-LWR}\bigl[k_0(s,s')\bigr]\,
\mathcal L_{s'-LWR}^{\! \top}
\end{equation}
Then, $K_{\mathrm{lin}}$ not only maintains positive definite but also is satisfied in the sense of generalization
$\mathcal L\,f=0$.

\subsection{Residual Kernel and Total Covariance}
Linearization discards all higher-order terms of $\mathcal O(\delta\rho^2)$, and at the same time
Observations also have measurement noise and local nonlinear effects. Introduce a "small-scale" SE residual kernel:
\begin{equation}
    K_{\mathrm{res}}(s,s')
  := B_{\mathrm{res-LWR}}\,
     \sigma_{\mathrm{res}}^2
     \exp\!\Bigl[-\frac{(x-x')^2}{2\tilde\ell_x^2}
                 -\frac{(t-t')^2}{2\tilde\ell_t^2}\Bigr]
\end{equation}
Due to the additive closure of the PD kernel, the final total covariance is defined as:
\begin{equation}\label{eq:lwr_total_cov_cn}
K_{\mathrm{tot-LWR}}(s,s')
= K_{\mathrm{lin-LWR}}(s,s')
+ K_{\mathrm{res}}(s,s').
\end{equation}

Unlike in the PEGP--ARZ model, where we introduce two Riemann invariants, the LWR model only contains one conservation equation for density. Therefore, only the scalar perturbation \(\delta\rho\) is used to establish the GP model, corresponding to the one-dimensional operator-embedded covariance kernel.

The posterior predictive distribution for the density perturbation is given by:
\begin{equation}
p\bigl(\delta\rho_* \mid \mathcal{D}\bigr)
\;\approx\;
\mathcal{N}\!\bigl(\mu_*,\,\Sigma_* + \sigma_n^2\bigr).
\end{equation}

To recover the original physical quantities, we map the perturbation back to density and velocity through:
\begin{align}
\rho_* &= \rho_0 + \delta\rho_*\,,\\
v_*    &\approx V(\rho_*) \approx v_0 + V'(\rho_0)\,\delta\rho_*\,.
\end{align}
These relationships can be expressed compactly in matrix form by defining the Joint mean and covariance as:
\begin{align}
F       &= \begin{pmatrix}
             1\\[4pt]
             V'(\rho_0)
           \end{pmatrix}\,,\\
\bm m_0 &= \begin{pmatrix}
             \rho_0\\[4pt]
             v_0
           \end{pmatrix}\,.
\end{align}
Under this linear mapping, the joint predictive distribution of density and velocity is therefore:
\begin{align}
    \begin{pmatrix}
\rho_*\\[4pt]
v_*
\end{pmatrix}
\;\sim\;
\mathcal{N}\!\Bigl(
\bm m_0 + F\,\mu_*,\;
F\,\bigl(\Sigma_* + \sigma_n^2\bigr)\,F^\top
\Bigr).
\end{align}

Finally, when incorporating the measurement noise on \(\rho\) and \( v \)   
with variances $\sigma_\rho^2$ and $\sigma_v^2$ respectively, the total predictive covariance becomes:
\begin{equation}
    \operatorname{Cov}\bigl[\rho_*,v_*\bigr]
= F\,(\Sigma_* + \sigma_n^2)\,F^\top
+ \operatorname{diag}(\sigma_\rho^2,\sigma_v^2).
\end{equation}

\end{document}